\documentclass[pdflatex,sn-mathphys-num]{sn-jnl}

\newcommand{\orcidlink}[1]{\href{https://orcid.org/#1}{\textcolor[HTML]{A6CE39}{\tiny (iD)}}}
\usepackage{graphicx}
\usepackage{multirow}
\usepackage{placeins}
\usepackage{amsmath}
\usepackage[numbers,sort&compress]{natbib}
\usepackage{svg}
\usepackage{mathrsfs}
\usepackage[title]{appendix}
\usepackage{xcolor}
\usepackage{textcomp}%
\usepackage{manyfoot}%
\usepackage{booktabs}%
\usepackage{longtable}
\usepackage{etoolbox}
\usepackage{algorithm}%
\usepackage{algorithmicx}%
\usepackage{algpseudocode}%
\usepackage{listings}%
\usepackage{subcaption}%

\theoremstyle{thmstyleone}%
\theoremstyle{thmstyletwo}%

\theoremstyle{thmstylethree}%

\robustify\bfseries

\begin{document}

\title[Article Title]{A Unified Multimodal Framework for Dataset Construction and Model-Based Diagnosis of Ameloblastoma}



\author[1]{\fnm{Ajo Babu} \sur{George} \orcidlink{0009-0005-3026-0959}}
\email{drajo\_george@dicemed.com}

\author[2]{\fnm{Anna Mariam} \sur{John} \orcidlink{0009-0003-1090-744X}}
\email{annamariamjohnn@gmail.com}
\equalcont{These authors contributed equally to this work.}

\author[3]{\fnm{Athul} \sur{Anoop} \orcidlink{0009-0005-8600-9050}}
\email{athulanoop2020@gmail.com}
\equalcont{These authors contributed equally to this work.}

\author[4]{\fnm{Balu} \sur{Bhasuran} \orcidlink{0000-0002-9890-4627}}
\email{balubhasuran08@gmail.com}
\equalcont{These authors contributed equally to this work.}


\affil[1]{\orgname{DiceMed}, \orgaddress{\city{Cuttack}, \country{India}}}

\affil[2]{\orgname{School of Sciences (SOS), Indira Gandhi National Open University}, \orgaddress{\city{New Delhi}, \state{New Delhi}, \country{India}}}

\affil[3]{\orgdiv{Computer Science Engineering}, \orgname{College of Engineering Trivandrum}, \orgaddress{\city{Thiruvananthapuram}, \state{Kerala}, \country{India}}}

\affil[4]{\orgname{School of Information, Florida State University}, \orgaddress{\city{Tallahassee}, \country{USA}}}


\abstract{Artificial intelligence (AI)-enabled diagnostics in maxillofacial pathology require structured, high-quality multimodal datasets. However, existing resources provide limited ameloblastoma coverage and lack the format consistency needed for direct model training. A newly curated multimodal dataset, specifically focused on ameloblastoma, has been developed. This dataset integrates annotated radiological, histopathological, and intraoral clinical images with structured data derived from case reports. Natural language processing techniques were employed to extract clinically relevant features from textual reports, while image data underwent domain-specific preprocessing and augmentation to improve diversity and representativeness.

Using this dataset, a multimodal deep learning model was developed to classify ameloblastoma variants, assess behavioral patterns such as recurrence risk and anatomical localization, and support surgical planning. Although electronic health record (EHR) data were not part of the training dataset, the model is designed to accept structured clinical inputs—such as presenting complaint, patient age, and gender—during deployment to enhance personalized inference. The model also provides treatment recommendations and outcome predictions upon analyzing previous ameloblastoma-related cases. 

Quantitative performance evaluation of the multi-label classification model demonstrated substantial improvements with data preprocessing. For example, variant classification accuracy increased from 46.2\% to 65.9\%, and abnormal tissue detection F1-score improved from 43.0\% to 90.3\%. Across all evaluated metrics, the model exhibited reduced variability and higher robustness, with significant gains in both absolute and relative performance. Statistical tests confirmed the significance of variant classification and abnormality detection improvements 

Performance was benchmarked against existing resources, including MultiCaRe, and validated through expert review to ensure clinical relevance. This work advances patient-specific decision support for ameloblastoma by providing both a robust dataset and an adaptable multimodal AI framework.
}




\maketitle
\section{Introduction}\label{sec1}

Artificial intelligence offers transformative potential for improving diagnostic accuracy and treatment planning in maxillofacial pathology. Comprehensive patient assessment in the field of maxillofacial diagnostics relies on integrating diverse data modalities, including radiological imaging, histopathology, and clinical observations. Nevertheless, the widespread application of AI is significantly impeded by the critical absence of large-scale, publicly accessible multimodal datasets. Proprietary and privacy concerns severely restrict access to detailed  electronic health record (EHR) data, aggravating this data scarcity \cite{unni2024automatic} and hindering the development of sophisticated AI models capable of capturing the subtle complexities of maxillofacial conditions, particularly underrepresented pathologies.

Addressing a fundamental limitation in the field, a novel framework is presented for the curation of comprehensive multimodal datasets from open-sourced case reports. The approach focuses specifically on ameloblastoma as a representative pathology, integrating annotated radiological, histopathological, and clinical images with structured textual data. Focusing specifically on ameloblastoma as a representative pathology, the approach integrates annotated radiological, histopathological, and clinical images with structured textual data. Advanced natural language processing techniques, including state-of-the-art language models, are leveraged to accurately extract clinically relevant information from unstructured case narratives. Utilizing this uniquely curated multimodal dataset, a deep learning model is developed aimed at improving the classification of ameloblastoma variants, predicting disease behavior, and ultimately enhancing clinical decision support and surgical planning \cite{zhou2017deep}. The methodologies developed herein are designed for broader applicability, providing a robust and extendable paradigm for advancing patient-specific AI applications in maxillofacial healthcare.

\section{Dataset}\label{sec2}

\subsection{Image Dataset Creation Pipeline}\label{subsec1}

\begin{figure}[H]
\centering
\includegraphics[width=0.9\textwidth]{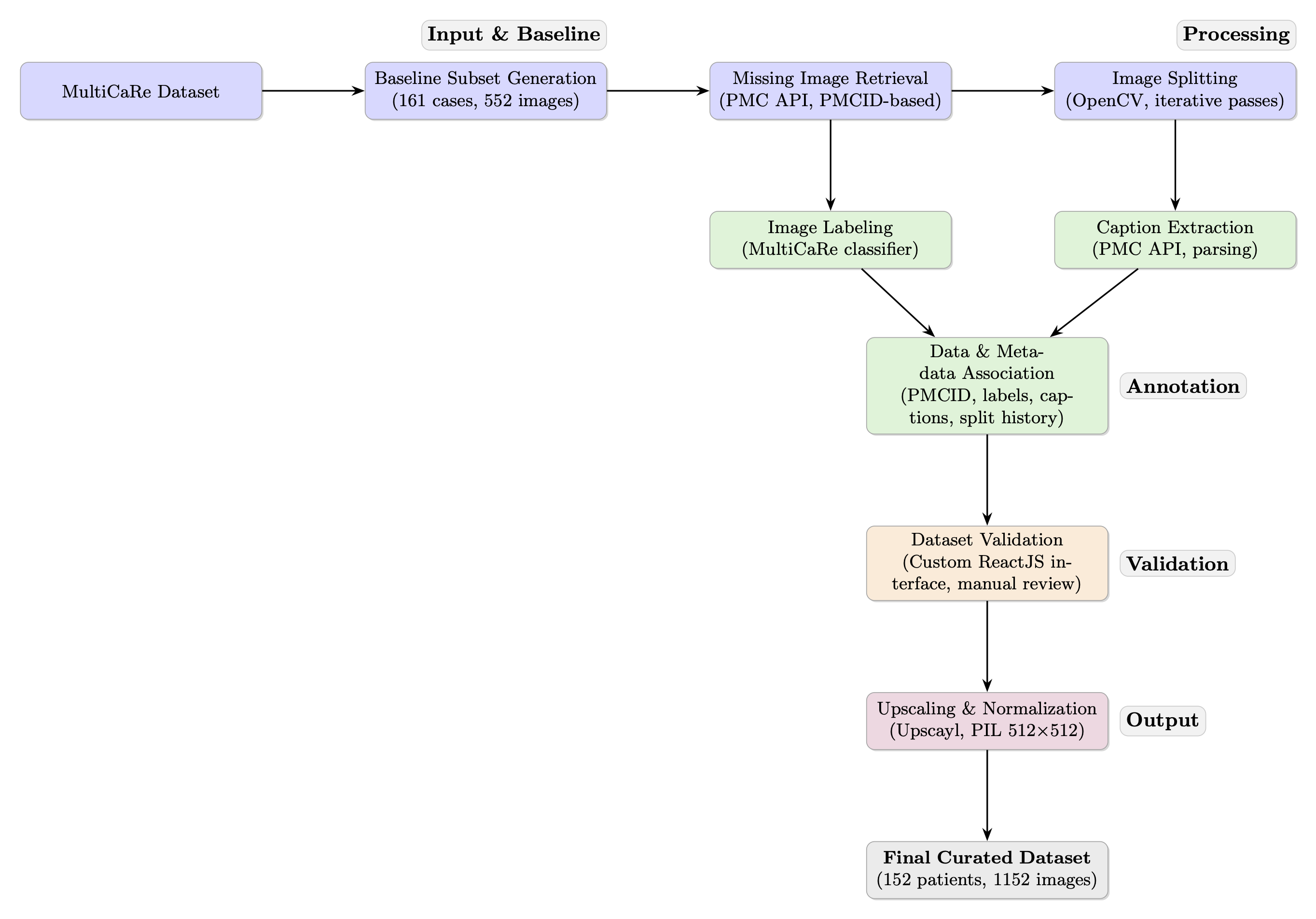}
\caption{Flowchart depicting the image dataset creation pipeline}\label{fig:fig1}
\end{figure}

To facilitate the development of a high-quality, structured dataset specific to ameloblastoma, the Python API provided by the MultiCaRe dataset~\cite{nievas2025multicaredataset} was utilized to generate a domain-specific subset. The resulting dataset was curated through a multi-step pipeline designed to address both quality control and data augmentation, as outlined in Fig.~\ref{fig:fig1}.

\subsubsection{Baseline Dataset Generation}\label{subsec:baseline-generation}

The baseline dataset comprised 161 unique patient case reports, yielding a total of 552 images. The gender distribution across patient data was balanced. However, several limitations were identified:

\begin{itemize}
    \item A portion of case reports lacked associated images entirely or included only partial imaging data.
    \item Multiple clinical images were erroneously annotated as pathology images.
    \item Some entries in the generated subset were not case reports, thereby introducing noise.
\end{itemize}

\subsubsection{Missing Image Retrieval}\label{subsec:missing-images}

Given the low image-to-patient ratio, it was hypothesized that many case reports lacked associated images. This hypothesis was subsequently validated through the analysis of a random sample of case reports, which confirmed incomplete image retrieval. To address this issue, the PMCID identifiers from the subset were utilized to query the PubMed Central (PMC) API, enabling the independent retrieval of missing images from the original publications.

\subsubsection{Image Splitting}\label{subsec:image-splitting}

Many retrieved images contained composite figures comprising multiple sub-images. To improve granularity, these composite images were split using an OpenCV-based approach, inspired by the methodology in the MultiCaRe paper\cite{multicare-dataset-paper}. However, the implementation involved two iterative passes to account for irregular layouts and ensure more comprehensive splitting. While this approach increased image resolution and modularity, it occasionally resulted in over-splitting and the inclusion of blank or partially informative image segments.

\subsubsection{Image Labeling}\label{subsec:image-labeling}

The classification of images into clinical, pathology, and radiology categories was performed using the pre-trained MultiCaRe classifier model. Labeling was conducted post-splitting to improve granularity and model performance. Nonetheless, several limitations of the classifier were evident:

\begin{itemize}
    \item Clinical images were occasionally mislabeled as pathology or radiological images.
    \item Disambiguation between similar imaging modalities (e.g., MRI vs. X-ray) was inconsistent.
\end{itemize}

These labeling inconsistencies are attributed to the classifier model itself rather than the pipeline design.

\subsubsection{Caption Extraction}\label{subsec:caption-extraction}

The MultiCaRe caption extraction pipeline was partially adopted to associate descriptive text with images. Case reports were retrieved via the PMC API, and captions were parsed accordingly. A key limitation observed was the inability of the pipeline to associate correct captions with sub-images in composite figures. In such cases, a single caption was often erroneously applied to all sub-images, reducing the specificity of image-text association.

\subsubsection{Data Association}\label{subsec:data-association}

Each image’s metadata incorporated its origin (PMCID), split history, label classification, and caption text. This metadata framework ensured comprehensive traceability. All associations were anchored to the originating PMCID, establishing a consistent link between images and their source case reports.

\subsubsection{Dataset Validation}\label{subsec:dataset-validation}

A semi-automated validation framework was developed using a custom ReactJS web interface. The interface facilitated visual review and manual correction of the dataset. Key features included:

\begin{itemize}
    \item Retrieval of case reports via PMCID.
    \item Synchronized display of associated image data and metadata.
    \item Support for editing existing metadata, uploading new images with metadata, and deleting erroneous entries.
\end{itemize}

This interface enabled effective validation of all case reports, leading to the removal of nine invalid entries and refinement of metadata for mislabeled or misclassified images.

\subsubsection{Upscaling and Normalization}\label{subsec:normalization}

To enhance visual clarity and standardize image dimensions, a two-stage image processing step was applied:

\begin{itemize}
    \item \textbf{Upscaling:} The \texttt{upscayl} UNIX CLI was used to upscale all images by a factor of 4 using the default model, which provides high-quality results under a non-restrictive license.
    \item \textbf{Normalization:} Images were resized to 512$\times$512 pixels using the Python Imaging Library (PIL) to ensure uniformity across the dataset.
\end{itemize}

\subsubsection{Final Dataset Characteristics}\label{subsec:final-dataset}

Following curation, the final dataset comprises:

\begin{itemize}
    \item 152 unique patients (after removal of 9 invalidated cases).
    \item 1,152 images, each categorized and stored by image type.
\end{itemize}

\begin{longtable}{llr}
\caption{Summary of Image Modalities and Class Labels} \label{tab:summary} \\
\toprule
\textbf{Category} & \textbf{Sub-Category / Label} & \textbf{Count} \\
\midrule
\endfirsthead
\multicolumn{3}{c}%
{{\bfseries \tablename\ \thetable{} -- continued from previous page}} \\
\toprule
\textbf{Category} & \textbf{Sub-Category / Label} & \textbf{Count} \\
\midrule
\endhead
\bottomrule
\endfoot
\multicolumn{2}{l}{\bfseries Image Modality} \\
\midrule
 & Radiology & 412 \\
 & Medical Photograph & 381 \\
 & Pathology & 354 \\
 & Chart & 2 \\
\midrule \multicolumn{3}{l}{\bfseries Class Labels} \\ \midrule\multicolumn{2}{l}{\itshape Core Modalities} \\
 & Radiology & 548 \\
 & Pathology & 477 \\
 & Medical Photograph & 421 \\
\midrule\multicolumn{2}{l}{\itshape Radiology Scans} \\
 & CT & 258 \\
 & X Ray & 218 \\
 & Panoramic & 201 \\
 & MRI & 36 \\
 & OPG & 32 \\
 & Cone Beam & 30 \\
 & Angiography & 2 \\
 & Ultrasound & 1 \\
 & Pet & 1 \\
 & Cta & 1 \\
\midrule\multicolumn{2}{l}{\itshape Radiology Views} \\
 & Axial & 116 \\
 & Dental View & 81 \\
 & Sagittal & 72 \\
 & Frontal & 29 \\
 & Occlusal & 18 \\
 & Periapical & 5 \\
 & Oblique & 4 \\
 & Ultrasound View & 1 \\
 & Bone Window & 1 \\
 & Posteroanterior & 1 \\
\midrule\multicolumn{2}{l}{\itshape Pathology Stains} \\
 & H\&E & 358 \\
 & Immunostaining & 58 \\
 & Pas & 10 \\
 & Ihc & 9 \\
 & Congo Red & 3 \\
 & Gram & 2 \\
 & Masson Trichrome & 2 \\
 & Van Gieson & 2 \\
 & Papanicolaou & 2 \\
 & Alcian Blue & 2 \\
 & Fish & 1 \\
 & Giemsa & 1 \\
\midrule\multicolumn{2}{l}{\itshape Clinical Descriptors} \\
 & Head & 464 \\
 & Mass & 59 \\
 & Thorax & 13 \\
 & Abdomen & 8 \\
 & Pelvis & 6 \\
 & Neck & 6 \\
 & Contrast & 6 \\
 & Spin Echo & 3 \\
 & Ankle & 2 \\
 & Lower Limb & 2 \\
 & T2 & 2 \\
 & Malignant & 2 \\
 & T1 & 1 \\
\midrule\multicolumn{2}{l}{\itshape Image Types} \\
 & Other Medical Photograph & 235 \\
 & Oral Photograph & 155 \\
 & 3D & 10 \\
 & Skin Photograph & 9 \\
 & Chart & 2 \\
\midrule\multicolumn{2}{l}{\itshape Other Labels} \\
 & Axial Region & 2 \\
 & Body Part & 2 \\
 & Region Specific View & 2 \\
\bottomrule
\end{longtable}

\subsubsection{Limitations}\label{subsec:dataset-limitations}

Despite the extensive curation and validation process, the dataset exhibits certain limitations:

\begin{itemize}
    \item A subset of case reports lacks images in one or more categories (clinical, pathology, or radiology).
    \item Final label accuracy requires expert medical validation, which is supported via the developed web interface but remains a manual process.
\end{itemize}

\subsection{Text Dataset Creation}\label{subsec:data-extraction}
\subsubsection{Keyword-Based Pattern Matching
}\label{data}

\begin{figure}[h]
\centering
\resizebox{\textwidth}{!}{\includegraphics[]{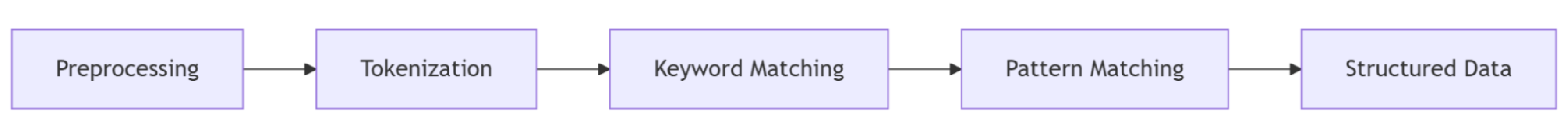}
}
\caption{Keyword Based Data Extraction Pipeline}\label{keyword}
\end{figure}

The keyword-based approach, as shown in Fig.~\ref{keyword}, employs a rule-driven pipeline to extract structured data from unstructured clinical text. After preprocessing and sentence segmentation, refined lists of medical terms and domain-specific keywords guide the identification of relevant entities. Regular expression patterns are applied to capture common linguistic structures. The method also accounts for relationships across sentences and includes fallback mechanisms to recover potentially missed information. All rules and keyword sets are grounded in clinical expertise, and case reports are processed sequentially to preserve clinical coherence.

\vspace{1em}

A comparable strategy was presented in \cite{bib13}, which described a rule-based system for structuring nephrology examination reports using specialized medical dictionaries and syntactic pattern recognition. Their findings highlight the critical role of accurate keyword selection, error-tolerant processing (e.g., typo correction), and attention to paragraph-level context in transforming free-text clinical documentation into structured formats.

\subsubsection{Word2Vec Semantic Approach
}\label{subsec:word2vec-semantic}

The Word2Vec-based method utilizes the pre-trained word2vec-google-news-300 model, which was trained on approximately 100 billion words from the Google News corpus \cite{MikolovTomas2013EEoW} using the Skip-Gram with Negative Sampling (SGNS) algorithm due to its effective capture of both syntactic and semantic regularities in biomedical text, thereby producing robust vector spaces\cite{BhasuranBalu2023Dakd}. Each word is represented as a 300-dimensional vector, and sentence embeddings are generated by averaging the vectors of constituent words. These embeddings are compared via cosine similarity against centroid vectors derived from predefined category keywords (e.g., clinical features, radiological findings, treatments). A similarity threshold of 0.65 is employed; if the similarity falls below this threshold or if out-of-vocabulary terms are encountered, the system reverts to traditional keyword matching. 

\vspace{1em}

\begin{figure}[h]
\centering
\resizebox{\textwidth}{!}{\includegraphics[]{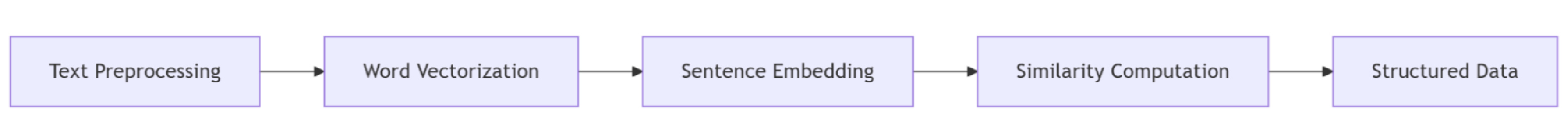}
}
\caption{Word2vec Based Data Extraction Pipeline}\label{word2vector}
\end{figure}

This hybrid approach balances semantic understanding with rule-based precision. However, managing large embedding models introduces computational overhead. Moreover, general-purpose embeddings like those from Google News may not adequately capture domain-specific nuances. Word embeddings trained in clinical corpora, such as electronic health records, have been shown to better represent medical terminology and improve performance in clinical information extraction tasks \cite{bib14}. This performance gap is a well-established concept in biomedical NLP, where domain-specific models significantly outperform their general-domain counterparts due to the specialized nature of the language \cite{BhasuranBalu2022BaSA}. Fig.~\ref{word2vector} depicts the working of the system.

\subsubsection{BioBERT Contextual Approach
}\label{subsec:word2vec}

\begin{figure}[h]
\centering
\resizebox{\textwidth}{!}{\includegraphics[]{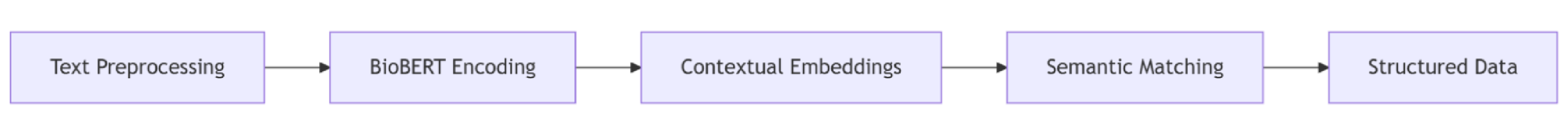}
}
\caption{Biobert Data Extraction Pipeline}\label{biobert}
\end{figure}

The BioBERT approach utilizes domain-specific language understanding through contextual embeddings, as portrayed in Fig.~\ref{biobert}. The BioBERT model generates contextual representations for sentences to enhance clinical language comprehension. These embeddings are compared to curated domain-specific examples to identify relevant medical information. After standard text cleaning and tokenization, extracted content is classified into categories such as presenting complaints, clinical features, and radiological findings. A hybrid strategy combines BioBERT-based semantic extraction with keyword matching, using rule-based extraction as a fallback mechanism. The overall pipeline integrates semantic, structural, and heuristic techniques for information extraction from medical text. This hybrid methodology of combining neural language models with rule-based approaches for extracting structured information from narrative text aligns with successful applications in other domains, where similar combinations of contextual understanding and pattern-based extraction have proven effective for processing complex textual narratives \cite{genaiscript}.

\vspace{1em}

This methodology aligns with the findings in \cite{bib15}, who introduced BioBERT—a domain-specific language representation model pre-trained on large-scale biomedical corpora. Their research demonstrated that BioBERT significantly outperforms BERT and previous state-of-the-art models in various biomedical text mining tasks, including named entity recognition, relation extraction, and question answering, by effectively capturing complex biomedical terminology and context.

\subsubsection{Gemini API Approach
}\label{subsec:gemini}

\begin{figure}[h]
\centering
\resizebox{\textwidth}{!}{
\includegraphics[width=0.9\textwidth]{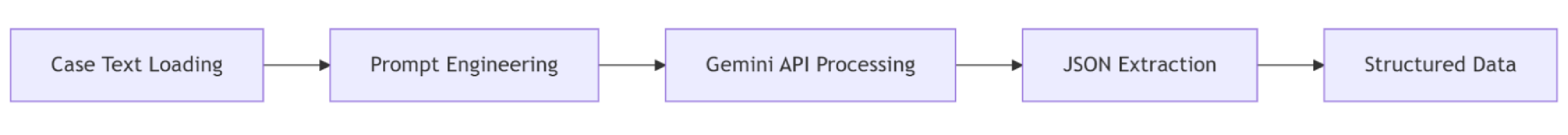}
}
\caption{GEMINI API Based Data Extraction Pipeline}\label{geminiapi}
\end{figure}

The Gemini approach, illustrated in Fig.~\ref{geminiapi}, employs a large language model guided by structured prompts to extract clinically relevant data from ameloblastoma case reports. Case texts are processed using detailed instructions that specify the exact information to be retrieved, including presenting complaints, clinical and radiological features, histopathology, treatment, and diagnosis. Responses are returned in a structured JSON format, organized by PMCID, with robust error handling and strict rules to avoid hallucinated outputs. The prompts guide the model to identify standard diagnostic categories—Solid/Multicystic, Unicystic, Peripheral, and Desmoplastic Ameloblastoma, ensuring medically accurate and consistent data extraction.

\FloatBarrier

\begin{figure}[h]
\centering
\includegraphics[width=\linewidth]{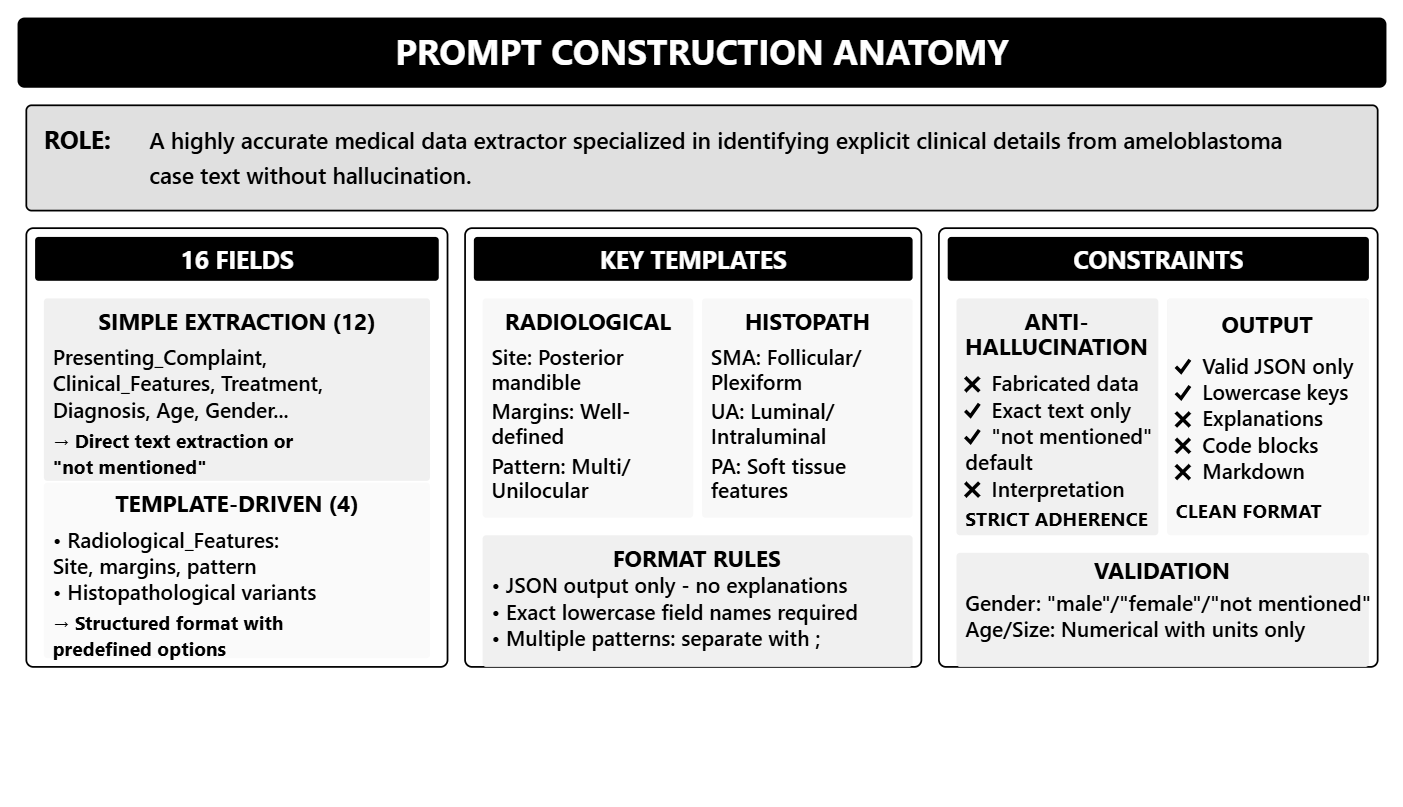}
\caption{GEMINI API Prompt Structure}
\label{geminipromptstructure}
\end{figure}

A recent study by \cite{AgrawalMonica2022LLMa} demonstrated that large language models can reliably extract structured information from clinical texts using carefully designed prompts. Building on this insight, the prompt was iteratively refined through trial and error to minimize hallucinations and improve clarity, ultimately producing reliable outputs verified by a medical expert. Importantly, the prompt was designed not just to match a predefined schema but to ensure the extraction of clinically relevant and accurate diagnostic and treatment details from authentic medical cases as shown in Fig. ~\ref{geminipromptstructure}. This demonstrates the potential of prompt-based LLM workflows, such as the one implemented with Gemini, to perform reliable biomedical data extraction without the need for task-specific model training. Comparable strategies integrating NLP with structured representations have also been explored in other domains, such as scam detection in social networks using graph neural networks \cite{syam2025graph}.

\FloatBarrier

Fig~\ref{geminipromptoutput} below displays the structured output of a particular PMCID obtained by Gemini.

\begin{figure}[h]
\centering
\includegraphics[width=0.8\textwidth]{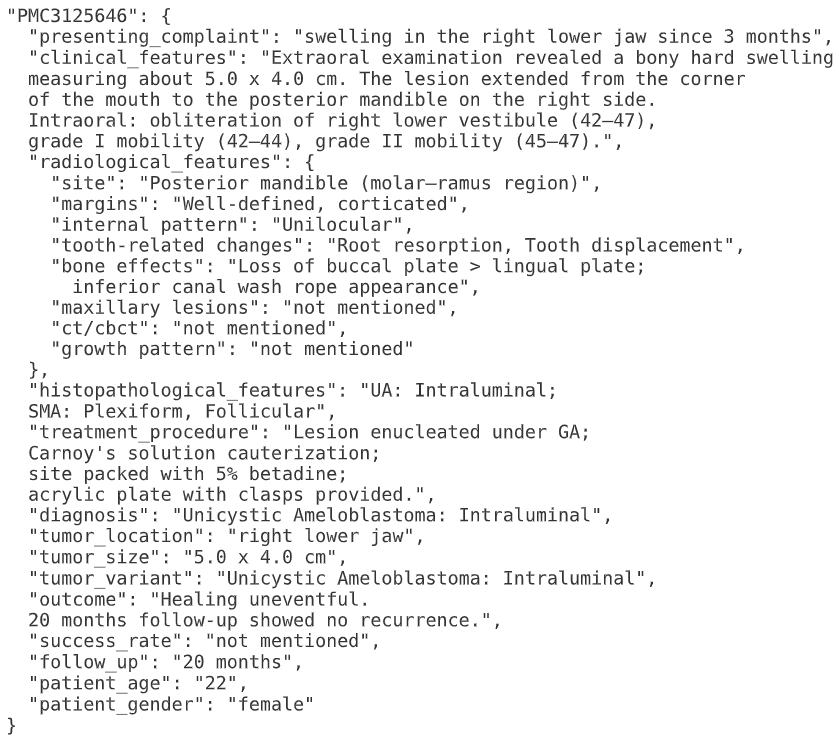}
\caption{GEMINI Output}
\label{geminipromptoutput}
\end{figure}
\FloatBarrier

\subsubsection{Comparative Advantages of Gemini Approach}\label{subsec:comparitive}

The Gemini approach offers notable advantages, particularly in capturing medical context through detailed prompting and producing clean, structured outputs. Its ability to handle missing data and generate consistent responses aligns with recent research indicating that instruction-tuned LLMs perform well in specialized domains like healthcare \cite{llm_paper}.

In comparison, rule-based and Word2Vec methods rely on fixed keywords and similarity thresholds, limiting their adaptability to complex clinical language. While BioBERT provides contextual understanding, it lacks the generative flexibility of models like Gemini. Overall, the Gemini-based approach proves more scalable and accurate for extracting information from complex medical reports, reinforcing the widely accepted view that LLMs, with effective prompting strategies, are well-suited for real-world, unstructured clinical data \cite{llm_paper}, a finding supported by their successful application in other clinical interpretation tasks \cite{HeZhe2024QoAo}.

\vspace{1em}

\section{Case-Based Retrieval System for Clinical Decision Support}

A case-based retrieval system was designed to match a patient's presenting complaints and clinical features with similar documented cases to support diagnosis and treatment planning for ameloblastoma.

\subsection{System Architecture}

The case-based retrieval system utilizes the structured data extraction methodologies described in Section~\ref{subsec:data-extraction} to enable clinicians to query the dataset using either free-text descriptions or structured form inputs. The system architecture consists of two main components:

\begin{itemize}
    \item \textbf{Text representation and embedding generation}
    \item \textbf{Similarity computation and case retrieval}
\end{itemize}

\subsection{Text Embedding Methods}
\label{subsec:embedding}
To enable efficient semantic matching between input queries and existing cases, several embedding approaches were implemented and compared:

\subsubsection{K-Nearest Neighbors with TF-IDF}

\begin{figure}[H]
\centering
\includegraphics[scale=0.28]{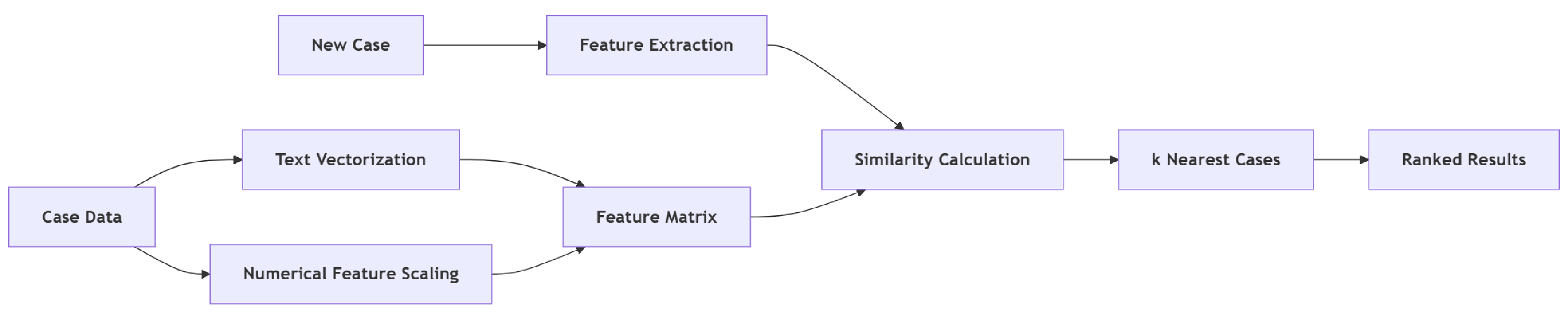}
\caption{KNN based Case Retrieval Pipeline}\label{knn_pipeline}
\end{figure}

The initial approach employed TF-IDF (Term Frequency-Inverse Document Frequency) vectorization to convert the textual case descriptions into numerical representations. A preprocessing pipeline was implemented to handle both textual features (e.g., presenting complaints, clinical features, histopathological descriptions) and numerical features (e.g., tumor size in mm). The pipeline is displayed in Fig.~\ref{knn_pipeline}.

\begin{itemize}
    \item Text normalization through stopword removal and stemming
    \item TF-IDF vectorization with a maximum of 500 features
    \item Cosine similarity metrics for nearest neighbor computation
\end{itemize}

\FloatBarrier

While this approach demonstrated reasonable performance for exact keyword matches, it lacked semantic understanding, resulting in suboptimal performance for cases described using different but semantically equivalent terminology. A similar use of TF-IDF combined with k-nearest neighbor search for identifying textual similarity was demonstrated by \cite{BithelShivangi2021UIoR}, where cosine distance over TF-IDF vectors was employed to retrieve semantically close documents in a legal case retrieval setting.

\subsubsection{K-Means Clustering}

To improve retrieval efficiency and better organize the case repository, a K-Means clustering approach was implemented. The preprocessing followed similar steps to the KNN approach, which is commonly used in similarity-based retrieval systems due to its effectiveness in nearest-neighbor search \cite{altman1992introduction}, but included additional components:

\begin{itemize}
    \item Feature standardization for numerical values
    \item Automated determination of optimal cluster count using the elbow method
\end{itemize}

\begin{figure}[H]
\centering
\includegraphics[width=0.9\textwidth]{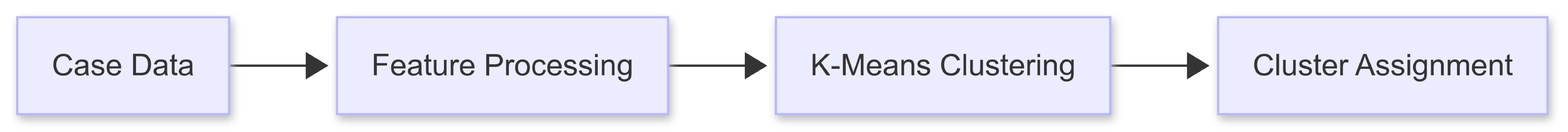}
\caption{K-Means based Text Retrieval Pipeline[Back-End]}\label{kmeans1}
\end{figure}

\begin{figure}[H]
\centering
\includegraphics[width=0.9\textwidth]{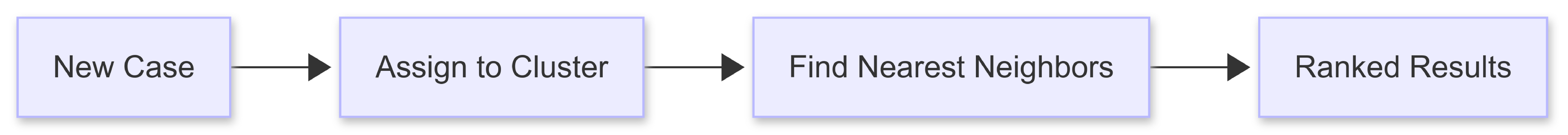}
\caption{K-Means based Text Retrieval Pipeline[User POV]}\label{kmeans2}
\end{figure}

The clustering approach, illustrated in Fig.~\ref{kmeans1}, organized cases into clinically meaningful groups based on feature similarity. This method improved retrieval speed by first identifying the relevant cluster for a new case and then searching within that cluster for the most similar cases as in Fig.~\ref{kmeans2}. However, the fixed nature of cluster boundaries sometimes resulted in suboptimal matches when cases fell near the cluster boundaries.

\subsubsection{Sentence-BERT with FAISS}
\label{subsec:sb-faiss}

To enable the semantically aware retrieval of case descriptions, a hybrid approach was adopted that integrates Sentence-BERT with FAISS (Facebook AI Similarity Search). Sentence-BERT, a derivative of BERT that uses a Siamese network architecture, was used to convert case texts into high-dimensional vectors capturing semantic meaning beyond simple word matching \cite{reimers2019sbert}. These embeddings were indexed using FAISS, a vector search library optimized for large-scale similarity search, facilitating efficient retrieval based on semantic proximity \cite{ghadekar2023sentence}.
\\ \\
This combined methodology offered the following key advantages:

\begin{itemize}
    \item Contextual understanding of medical terminology
    \item Capture of semantic relationships between different phrasings
    \item Significantly improved retrieval performance for cases with semantically similar descriptions
    \item Efficient similarity search, even with large case repositories
\end{itemize}

The Fig.~\ref{sentenceb} below shows the proposed combined methodology:

\begin{figure}[H]
\centering
\includegraphics[scale=0.266]{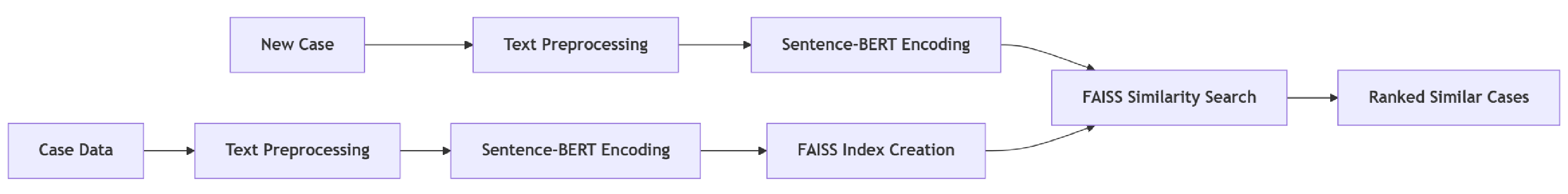}
\caption{Pipeline for Sentence Similarity-based Case Retrieval using S-BERT and FAISS}\label{sentenceb}
\end{figure}

The \texttt{all-MiniLM-L6-v2} lightweight variant of Sentence-BERT was selected for its optimal balance between computational efficiency and semantic accuracy. It  generates 384-dimensional embeddings that are well-suited for real-time inference. Unlike traditional BERT models that require pairwise comparisons during inference, Sentence-BERT enables independent encoding of sentences, followed by cosine similarity matching, thereby reducing inference complexity from quadratic to linear \cite{reimers2019sbert}. Low-latency similarity searches were performed using FAISS’s \texttt{IndexFlatL2},  demonstrating scalability to thousands of embeddings and enabling real-time semantic retrieval in high-volume datasets.

\subsection{Sentence-BERT with FAISS based Methodology}

Building upon the comparative analysis, a comprehensive case retrieval pipeline was implemented using the Sentence-BERT with FAISS approach. This methodology integrates advanced semantic understanding with efficient similarity search capabilities, making it particularly well-suited for clinical decision support in ameloblastoma management. \\ \\ The methodology consists of two primary phases: 
\begin{itemize}
    \item Data preprocessing and \item Query processing.
\end{itemize}

\subsubsection{Data Preprocessing Phase}
\label{datapreprocess}
The preprocessing pipeline as shown in Fig.~\ref{pre_image} transforms heterogeneous case reports into standardized representations suitable for semantic encoding:

\begin{figure}[h]
\centering

\includegraphics[scale=0.16]{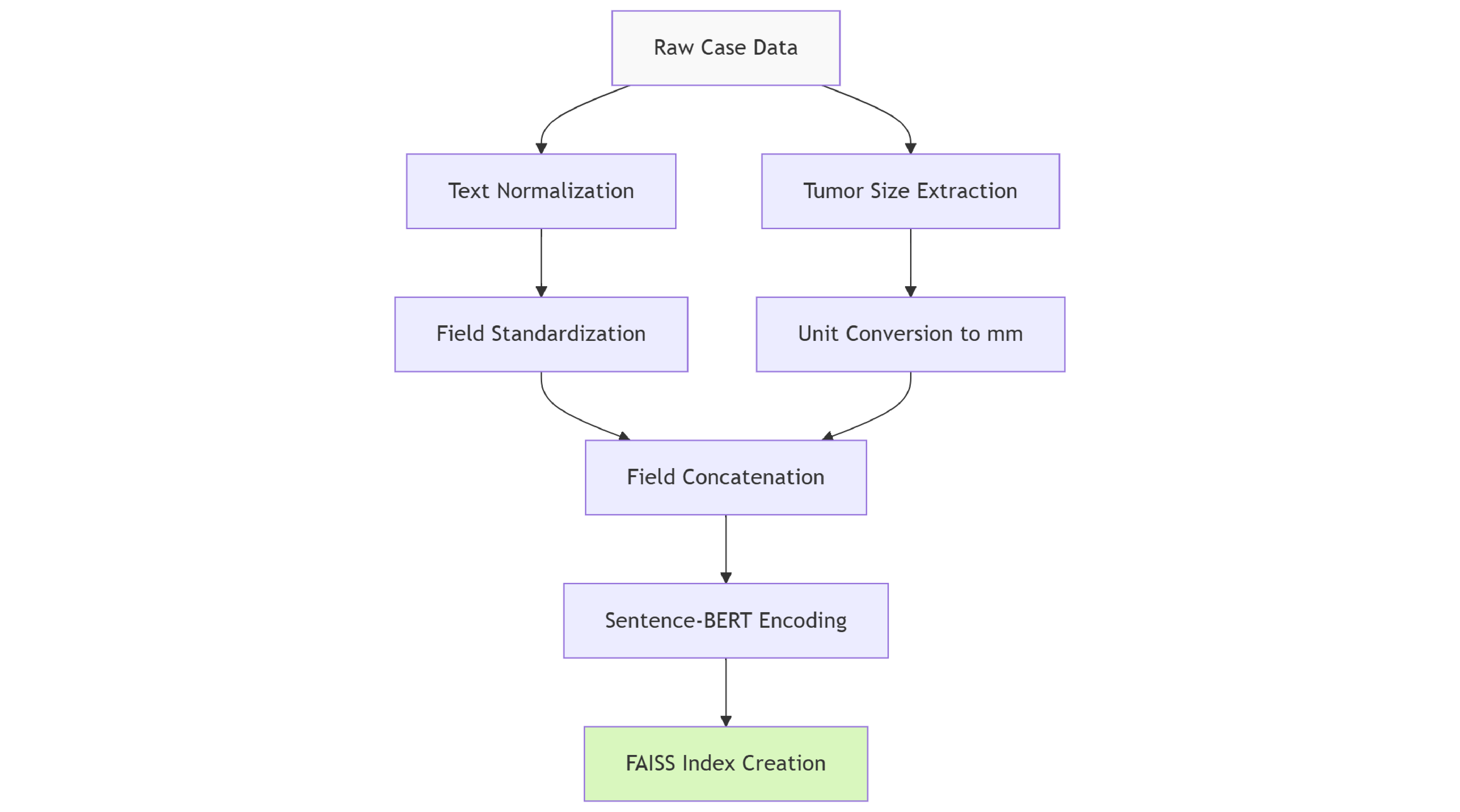}

\caption{Preprocessing Pipeline}\label{pre_image}
\end{figure}

\begin{itemize}
    \item \textbf{Text Normalization:} Clinical text undergoes cleaning to remove irrelevant content, standardize medical terminology, and correct inconsistencies in formatting.
    
    \item \textbf{Field Structuring:} Case reports are segmented into standardized fields (e.g., presenting complaints, clinical features, radiological features) using the techniques described in Section~\ref{subsec:data-extraction}.
    
    \item \textbf{Dimension Standardization:} Tumor size measurements are normalized to millimeters using specialized regex patterns that handle various expression formats (e.g., ``2 cm x 3 cm'', ``19.8 mm mesiodistally'').
    
    \item \textbf{Case Representation:} Processed fields are concatenated with explicit field markers to maintain semantic relationships while creating a unified text representation of each case as follows:
    \begin{quote}
\texttt{Presenting complaint: [text]. Clinical features: [text]. Radiological features: [text]. Histopathological features: [text]. Tumor location: [text]. Diagnosis: [text]. Tumor size: [text]. Tumor variant: [text]. Patient age: [text]. Patient gender: [text].}
\end{quote}
    
    \item \textbf{Semantic Encoding:} The \texttt{all-MiniLM-L6-v2} Sentence-BERT model converts case representations into 384-dimensional dense vector embeddings, capturing the semantic essence of each case report.
    
    \item \textbf{FAISS Indexing:} Vector embeddings are organized into a FAISS \texttt{IndexFlatL2} structure, enabling efficient approximate nearest neighbor search with logarithmic time complexity.
\end{itemize}

\subsubsection{Query Processing Phase}
\label{query processing}
When a clinician submits a new case, the query processing phase executes the following steps:

\begin{itemize}
   
    \item \textbf{Query Preparation:} User input (free text or structured form) is processed using identical preprocessing steps to ensure consistency with indexed cases.
    
    \item \textbf{Query Encoding:} The same Sentence-BERT model generates a vector embedding for the query, placing it in the same semantic space as the case repository.
    
    \item \textbf{Similarity Search:} The FAISS index performs an efficient search to identify the most similar case vectors based on L2 distance.
    
    \item \textbf{Results Ranking:} Retrieved cases are ranked by similarity score and presented to the clinician with relevant clinical details and reference information.
\end{itemize}

A combined methodology incorporating Data Preprocessing and Query Processing can be seen in Fig.~\ref{overall_meth} below:

\begin{figure}[h]
\centering
\includegraphics[scale=0.266]{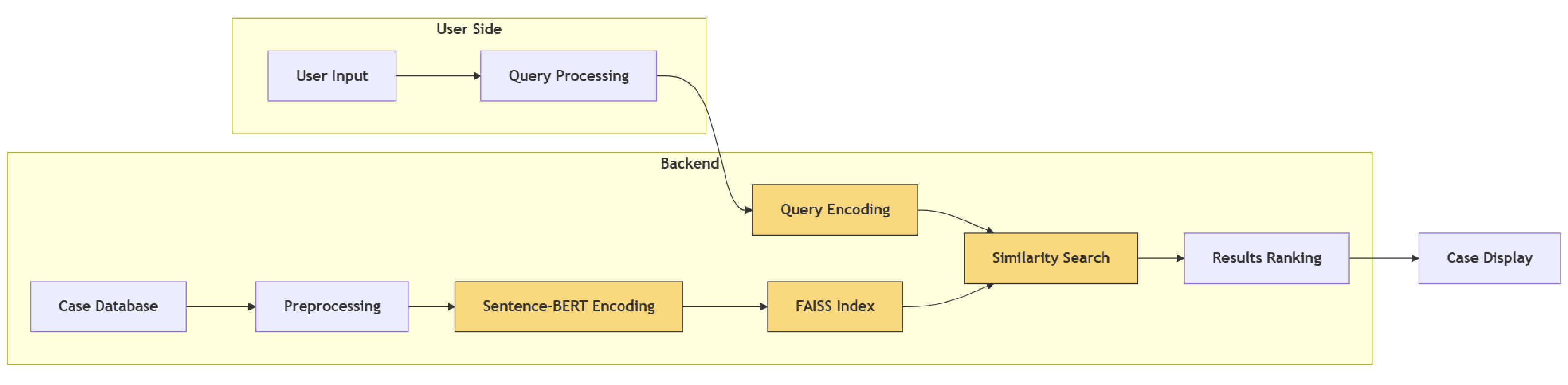}
\caption{Pipeline for Retrieval of Similar Clinical Cases}\label{overall_meth}
\end{figure}
\FloatBarrier

This methodology addresses the limitations of traditional keyword-based approaches by utilizing the semantic understanding capabilities of Sentence-BERT. By representing cases and queries in a shared semantic vector space, the system can identify conceptually similar cases even when different terminology is used to describe the same clinical features.

\subsection{Performance Considerations}
Each method was assessed based on multiple factors:

\begin{itemize}
    \item \textbf{Semantic Understanding:} The ability to match conceptually similar cases even when different terminology is used.
    \item \textbf{Retrieval Speed:} The computational efficiency, particularly for interactive use.
    \item \textbf{Implementation Complexity:} The ease of deployment and maintenance.
    \item \textbf{Scalability:} The ability to handle growing case repositories.
\end{itemize}

\subsubsection{Performance Comparison}

A benchmark evaluation was performed on multiple similarity search methods using a standardized dataset of medical cases. Key textual fields such as complaints, clinical features, radiological findings, and diagnoses were preprocessed into uniform representations for fair comparison.\\
\\
\textbf{Evaluation Framework:}
\begin{itemize}
    \item Identical query set applied to all methods
    \item Measured metrics included total execution time, vectorization/encoding time, query response time, memory consumption, and CPU utilization
    \item Computational complexity analyzed for both time and space efficiency
\end{itemize}

\begin{table}[h]
\centering
\small 
\caption{Performance metrics of the evaluated methods}
\label{comparison}
\begin{tabular}{lcc}
\hline
 & KNN with TF-IDF & K-Means Clustering \\
\hline
Total Time (s)       & 0.753 & 1.225 \\
Vectorization (s)    & 0.122 & 0.124 \\
Query (s)            & 0.105 & 0.102 \\
CPU (\%)             & 24.13 & 21.36 \\
\hline
\end{tabular}

\vspace{0.5em}
\parbox{0.95\linewidth}{\small \textit{Note:} KNN complexity is $O(n)$ query and $O(n \cdot d)$ space. K-Means is $O(k)$ query with $O(nki)$ training complexity.}
\end{table}

\subsubsection{Comparative Analysis and Final Selection}
The implemented methods showed distinct performance characteristics in the evaluation, as shown in Table~\ref{comparison}. KNN with TF-IDF offered simplicity and low memory usage but struggled with semantic understanding and exhibited poor scalability with its O(n) query complexity. K-Means improved organization through clustering with moderate CPU utilization (21.36\%), but its rigid cluster boundaries often misclassified borderline cases.

\vspace{0.5em}

Sentence-BERT with FAISS emerged as the optimal approach for the ameloblastoma case retrieval system by addressing these limitations. Its dense vector representations capture semantic meaning rather than just keyword matching which is critical for medical terminology where the same condition may be described differently across institutions. The approach's O(log n) query complexity ensures consistent performance as the database grows, while its semantic understanding enables matching of conceptually similar cases regardless of terminology variations. Despite a higher initial computational cost, this methodology provides superior retrieval quality and response time during clinical use, making it uniquely suited to support real-time decision-making in rare conditions such as ameloblastoma, where subtle presentation differences can significantly impact diagnosis and treatment planning.

 \subsection{Modern Embedding Baseline Validation}

\subsubsection{Evaluation Framework}
To validate this embedding approach against current standards in retrieval systems and address evolving semantic similarity methods, a comprehensive evaluation of modern embedding models was conducted. This evaluation addresses recent advances in dense retrieval systems and the availability of specialized biomedical language models that potentially offer superior performance for medical case matching, consistent with practices established in large-scale benchmarks such as the Massive Text Embedding Benchmark \cite{muennighoff2023mtebmassivetextembedding}.

\subsubsection{Evaluation Methodology}

The comparative assessment employed a standardized evaluation protocol to ensure fair comparison across diverse embedding architectures, following methodologies established in recent contrastive learning research \cite{gao2022simcsesimplecontrastivelearning}. All methods processed identical case representations using the preprocessing pipeline established in Section~\ref{datapreprocess}, maintaining consistency in text normalization, field structuring, and case formatting. This approach isolates embedding quality differences while controlling preprocessing variations that could confound results.

\paragraph{Query Selection}  
Five medical queries representing typical clinical scenarios were used:
\begin{itemize}
    \item Radiological presentations
    \item Histopathological feature matching
    \item Symptom-based case retrieval
    \item Tumor characteristic searches
    \item Complex multi-modal clinical cases
\end{itemize}
These queries were chosen to capture a broad range of medical terminology complexity and clinical relevance in ameloblastoma case management.

\vspace{3mm}

\paragraph{Similarity Computation}  
Cosine similarity was used to measure semantic alignment between query embeddings and case embeddings, and all embeddings were L2-normalized to ensure consistency. For each query, the top-5 most similar cases were retrieved and ranked based on similarity scores, normalized to the [0,1] range, where 1.0 indicates perfect alignment.

\vspace{3mm}

\paragraph{Performance Metrics}  
The following aspects of each method were evaluated:
\begin{itemize}
    \item \textbf{Initialization Time:} Time required to load the model and perform its first inference. This indicates how quickly a model can be deployed in a real-world clinical setting.
    \item \textbf {Model Size (MB):} Storage footprint of model parameters, derived analytically as(parameter count × 4 bytes) for FP32. Following the FAISS principle \cite{JohnsonJeff2021BSSw} , originally applied to index storage,this formulation is adapted here to model parameters. These values are determined solely by the architecture and remain independent of dataset size.

    \item \textbf{Similarity Quality:} Average cosine similarity across all queries, used to measure retrieval effectiveness. Higher scores indicate stronger semantic alignment between clinical queries and stored cases, directly supporting decision-making accuracy.
\end{itemize}

\subsubsection{Criteria for Model Selection}

The evaluation includes three categories of embedding approaches, selected to assess both general and domain-specific performance:
\begin{itemize}
    \item \textbf{General-Domain Models:}  
Sentence-BERT variants (all-MiniLM-L6-v2, all-mpnet-base-v2) representing widely-used general-purpose embeddings with strong performance across diverse text similarity tasks \cite{WangWenhui2020MDSD}.\\

\item\textbf{Modern Retrieval Systems:}  
Advanced embedding models such as BGE (BAAI/bge-base-en-v1.5) and E5 (intfloat/e5-base-v2), representing the current state-of-the-art in text embedding. For instance, E5-base achieves superior performance across diverse tasks, scoring 60.4 on average across 56 datasets in the MTEB benchmark while requiring significantly fewer parameters than competing models \cite{WangLiang2022TEbW}. E5 notably established a new milestone as the first model to outperform the traditional BM25 baseline on BEIR retrieval tasks using only unsupervised learning.\\

\item\textbf{Medical Domain Specialists:}  
ClinicalBERT \cite{HuangKexin2019CMCN} and PubMedBERT \cite{GuYu2022DLMP}, trained on clinical notes and biomedical literature respectively, providing domain-specific language understanding to evaluate the added value of medical specialization in clinical case retrieval.

\end{itemize}

\begin{figure}[H]
\centering
\includegraphics[width=0.7\textwidth]{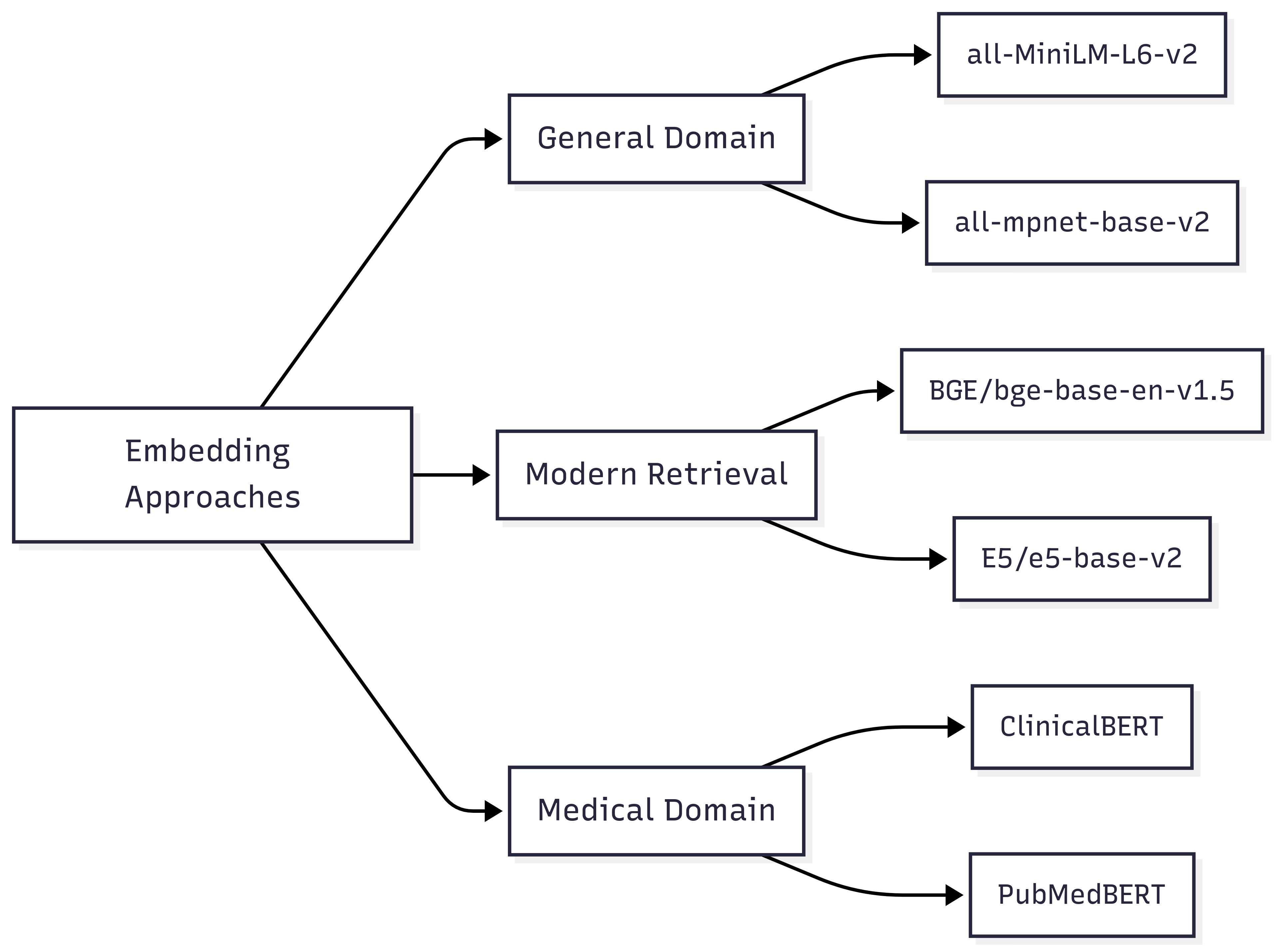}
\caption{Embedding Approaches}\label{kmeans1}
\end{figure}

\subsubsection{Performance Comparison and Analysis}

The evaluation of embedding models on the curated repository of 152 ameloblastoma cases revealed clear differences in both efficiency and retrieval effectiveness (Table~\ref{tab:embedding_performance}). The all-MiniLM-L6-v2 variant was the fastest to load, requiring only 3.9 seconds, and was also the most compact at 86 MB. This makes it particularly suitable for environments requiring rapid initialization or operating under real-time constraints. In contrast, all BERT-base architectures—including PubMedBERT, ClinicalBERT, BGE, and E5—ranged between 411–416 MB, reflecting a nearly five-fold increase in storage requirements. Such differences underscore the practicality of lightweight models in resource-limited clinical settings where hardware availability may be constrained.

Retrieval quality, however, showed the opposite trend. PubMedBERT achieved the highest similarity score (0.969), followed by ClinicalBERT (0.882), highlighting the advantages of domain-specific pretraining for biomedical applications \cite{alsentzer2019publiclyavailableclinicalbert,GuYu2022DLMP}. Modern retrieval models such as E5 (0.863) and BGE (0.772) offered a strong balance between general-purpose and medical-domain embeddings, aligning with recent benchmarks of contrastive embedding models \cite{WangWenhui2020MDSD,muennighoff2023mtebmassivetextembedding}. General-purpose models like MiniLM (0.630) and mpnet (0.646) produced lower similarity scores but still delivered clinically relevant matches. These findings are consistent with prior results on Sentence-BERT efficiency \cite{reimers2019sbert} and illustrate the well-documented trade-off between compactness and semantic accuracy \cite{muennighoff2023mtebmassivetextembedding}.

Overall, the results show a hierarchy across efficiency and retrieval accuracy. MiniLM stands out when responsiveness and storage efficiency are prioritized, while PubMedBERT provides the strongest semantic alignment at higher computational cost. Given that only 152 ameloblastoma cases were available, the smaller MiniLM model was chosen for deployment to ensure responsiveness and ease of integration, with the option to incorporate heavier, domain-specialized models in the future when greater retrieval precision is required.

\begin{table}[htbp]
\centering
\caption{Embedding models: initialization time, size, and similarity.}
\label{tab:embedding_performance}
\small
\setlength{\tabcolsep}{4pt} 
\renewcommand{\arraystretch}{1.2} 
\begin{tabular}{|p{0.28\textwidth}|p{0.22\textwidth}|p{0.13\textwidth}|p{0.14\textwidth}|p{0.14\textwidth}|}
\hline
\textbf{Model} & \textbf{Category} & \textbf{Init Time (s)} & \textbf{Model Size (MB)} & \textbf{Similarity Score} \\
\hline
\textbf{Sentence-BERT} \newline (all-MiniLM-L6-v2) & General & \textbf{3.9} & \textbf{86.2} & 0.630 \\
\hline
\textbf{Sentence-BERT} \newline (all-mpnet-base-v2) & General & 32.7 & 415.8 & 0.646 \\
\hline
\textbf{BGE} \newline (BAAI/bge-base-en-v1.5) & Modern Retrieval & 39.3 & 415.8 & 0.772 \\
\hline
\textbf{E5} \newline (intfloat/e5-base-v2) & Modern Retrieval & 39.0 & 415.8 & 0.863 \\
\hline
\textbf{ClinicalBERT} & Medical Domain & 38.9 & 411.3 & 0.882 \\
\hline
\textbf{PubMedBERT} & Medical Domain & 33.6 & 415.8 & \textbf{0.969} \\
\hline
\end{tabular}
\end{table}
\FloatBarrier

\subsubsection{Method Selection Validation}

Beyond raw performance, deployment feasibility was a decisive factor in model choice (Table~\ref{tab:embedding_performance}). While PubMedBERT delivered the highest semantic similarity, its substantially larger size and initialization overhead limit practicality in constrained clinical environments. MiniLM, by contrast, offers near-instant startup and minimal storage requirements, making it more suitable for real-time decision support where efficiency and responsiveness are essential \cite{muennighoff2023mtebmassivetextembedding}.  

The choice of MiniLM is further supported by the size and scope of the dataset: with 152 ameloblastoma cases, a lightweight model provides adequate retrieval performance while ensuring efficient integration into real-world clinical workflows. This aligns with prior evidence that smaller models can be preferable in domain-specific, limited-data contexts \cite{reimers2019sbert,HuangKexin2019CMCN}. Importantly, the modular architecture of the retrieval system allows higher-capacity embeddings such as PubMedBERT to be substituted in future, once larger case repositories become available and greater semantic precision can be justified. This  ensures that deployment feasibility is not sacrificed while maintaining compatibility with advances in biomedical embedding research \cite{GuYu2022DLMP,WangLiang2022TEbW}.

\subsection{Clinical Use Case Scenario}

The following use case demonstrates the practical application of the case-based retrieval system in a clinical setting; Fig.~\ref{systemflow} below displays its flow.

\begin{figure}[h]
\centering
\resizebox{\textwidth}{!}{\includegraphics[]{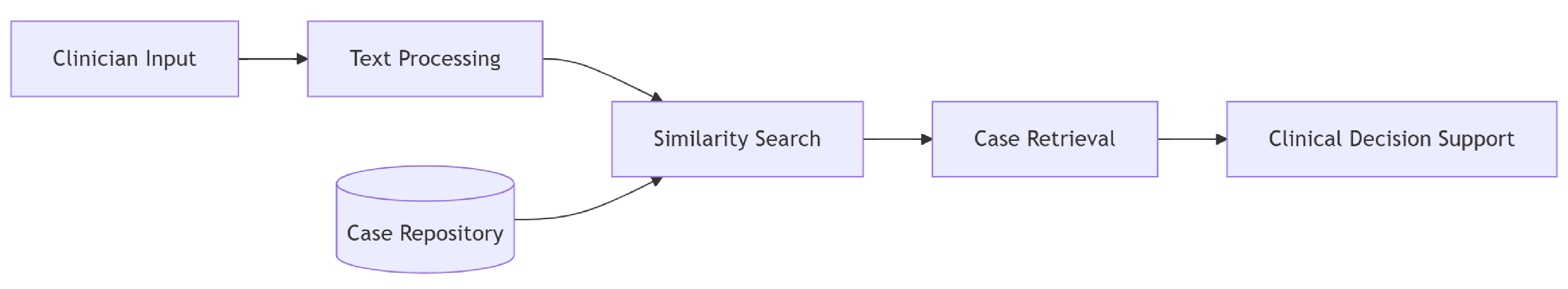}}
\caption{System Flow}\label{systemflow}
\end{figure}

\subsubsection{Patient Presentation}

A 43-year-old female patient presents with a gradually enlarging painless swelling in the right mandibular region that has been present for approximately 8 months. Clinical examination reveals a firm, non-tender swelling with intact overlying mucosa. Panoramic radiography shows a multilocular radiolucent lesion extending from the right mandibular first molar to the ramus.

\subsubsection{System Query and Retrieval}

The clinician inputs the information into the system using the free-text query mode:

\begin{quote}
\texttt{Painless swelling in right mandible, gradually enlarging over 8 months. Firm, non-tender with intact mucosa. Radiograph shows multilocular radiolucent lesion from first molar to ramus.}
\end{quote}

The Sentence-BERT with FAISS retrieval system processes this query and identifies semantically similar cases from the repository, returning the top matches ranked by similarity score.

\subsubsection{System Output and Clinical Decision Support}

The system presents several similar cases with the following key information:

\vspace{0.5em}
\noindent\textbf{Most similar case (Similarity score: 0.89)}
\begin{itemize}
    \item \textbf{Diagnosis:} Follicular ameloblastoma
    \item \textbf{Treatment:} Segmental mandibulectomy with fibular free flap reconstruction
    \item \textbf{Tumor size:} 4.5 × 3.2 cm
    \item \textbf{Patient demographics:} 47-year-old male
    \item \textbf{Reference ID:} PMC7234567
\end{itemize}

\vspace{0.5em}
\noindent\textbf{Second-ranked case (Similarity score: 0.85)}
\begin{itemize}
    \item \textbf{Diagnosis:} Plexiform ameloblastoma
    \item \textbf{Treatment:} Partial mandibulectomy with reconstruction plate
    \item \textbf{Tumor size:} 3.8 × 2.5 cm
    \item \textbf{Patient demographics:} 39-year-old female
    \item \textbf{Reference ID:} PMC8456123
\end{itemize}

\vspace{0.5em}
\noindent\textbf{Third-ranked case (Similarity score: 0.81)}
\begin{itemize}
    \item \textbf{Diagnosis:} Follicular ameloblastoma
    \item \textbf{Treatment:} Marginal mandibulectomy with iliac crest bone graft
    \item \textbf{Tumor size:} 3.2 × 2.8 cm
    \item \textbf{Patient demographics:} 52-year-old female
    \item \textbf{Reference ID:} PMC6543210
\end{itemize}

\subsection{Implementation Challenges and Solutions}

The development of the case-based retrieval system presented several technical challenges that required innovative solutions, particularly in the data extraction, text representation, and computational efficiency domains.

\subsubsection{Data Structuring Challenges}

A primary challenge was the effective structuring of heterogeneous clinical data. Case reports varied significantly in their descriptions of histopathological characteristics, radiological findings, and clinical presentations, making direct comparison difficult. This issue was addressed by the implementation of a system that standardized field names, developed specialized regex patterns for normalizing measurements, and created a unified case representation format. This structured approach significantly improved match accuracy by ensuring consistent representation across diverse case reports.

\subsubsection{Computational Efficiency Challenges}

Computational efficiency was essential for developing an interactive clinical application with responsive performance. Early implementations using TF-IDF and K-Means were memory-intensive and required frequent retraining as new cases were added. To balance accuracy with practical constraints, various methods including dimensionality reduction and feature selection, were explored. Sentence-BERT combined with FAISS was ultimately adopted, offering fixed-size dense embeddings and efficient similarity search, thereby achieving a strong balance between semantic understanding and computational performance.

\subsubsection{Method Exploration and Fallback Mechanisms}

During the research, multiple data extraction techniques were evaluated for processing unstructured case reports. A cascading fallback mechanism was implemented for text extraction: beginning with Sentence-BERT semantic matching, falling back to Word2Vec when confidence was low, and defaulting to keyword-based pattern matching as a last resort. Similarly, the search pipeline prioritized FAISS-based similarity search, with fallbacks to KNN for short or sparse queries, and keyword matching when semantic methods underperformed. This hierarchical approach ensured robust information retrieval across varying input quality levels.

\vspace{1em}

\section{Multi-Task Deep Learning Framework for Histopathological Image Classification}

The proposed framework utilizes a multi-output \cite{aljuaid2024multi} convolutional neural network to simultaneously predict abnormality status, diagnosis classification, and tumor variant from histopathological images.

\subsection{Data Preparation Pipeline}

\begin{figure}[h]
\centering
\resizebox{\textwidth}{!}{
\includegraphics[]{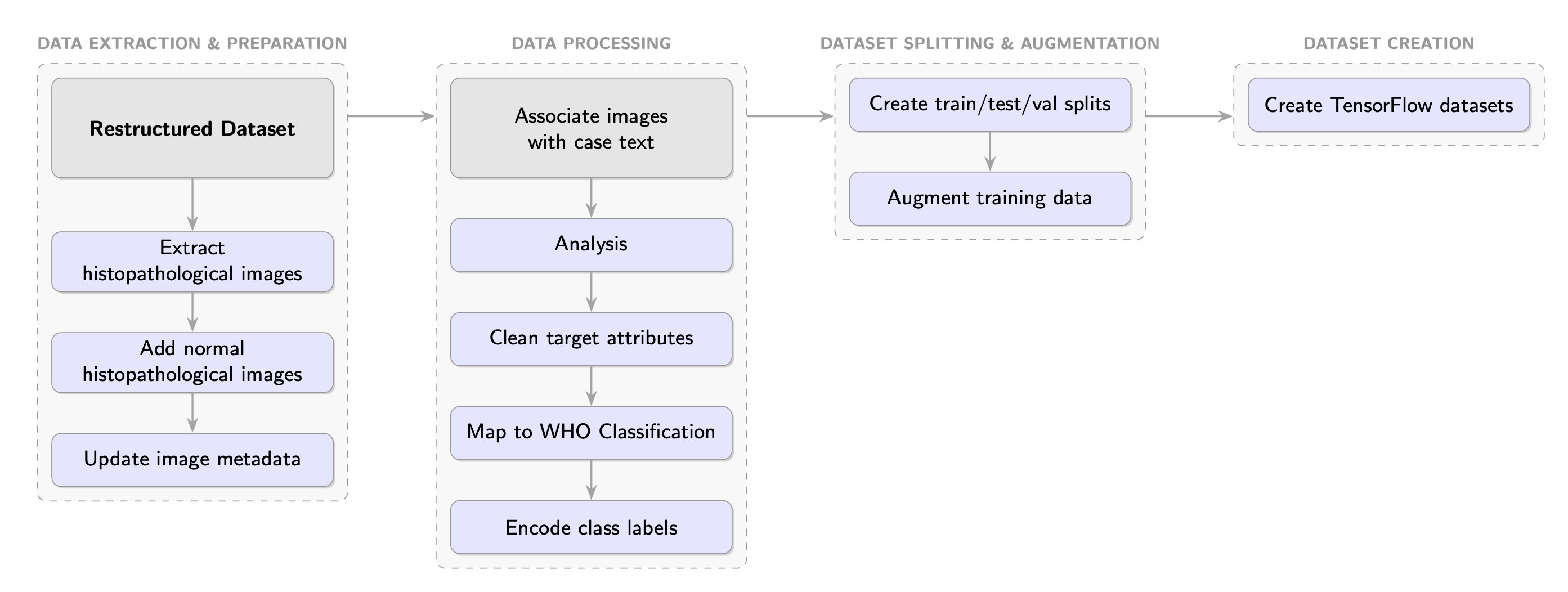}
}
\caption{Data Preparation Pipeline}
\label{fig:data_preparation_pipeline}
\end{figure}

The restructured dataset served as the primary source of histopathological images corresponding to documented cases of ameloblastoma. The initial phase of the data processing pipeline involves systematic data extraction and preparation, as illustrated in Fig.~\ref{fig:data_preparation_pipeline}. Histopathological images were selectively retrieved from the restructured dataset by leveraging the associated metadata labels to ensure accurate and case-specific extraction. To enhance the representativeness and diagnostic coverage of the dataset, typical histopathological images—sourced from publicly available repositories such as Kaggle~\cite{ahmed2023histopathologic} — were incorporated into the corpus. This augmentation was performed to ensure the dataset encapsulates typical patterns of ameloblastoma. Following image integration, the metadata was meticulously updated to reflect the augmented dataset structure, thereby facilitating downstream processing and analysis with enhanced contextual relevance.

\FloatBarrier

The second phase of the pipeline, encompasses the processing stage wherein relevant textual attributes—extracted from structured case reports—are systematically associated with the corresponding histopathological images using the unique PMCID as a reference identifier. During this stage, the extracted textual information is integrated into the existing metadata schema. Additionally, a preliminary surface-level analysis of the frequency distributions of key attributes was conducted to gain insights into data characteristics and to inform the selection of appropriate preprocessing strategies.

Based on this analysis, targeted cleaning and normalization procedures were devised. Given the categorical nature of the target attributes (e.g., diagnosis, tumor variant), it was deemed necessary to transform these attributes into a standardized and uniform taxonomy of class labels. The class nomenclature was curated in alignment with the World Health Organization (WHO) 2022 Classification System \cite{2022-who-classification-system}, ensuring clinical and pathological validity.

To achieve accurate mapping from unstructured text to standardized class labels, a hybrid approach combining rule-based logic and fuzzy string matching was employed. This approach facilitated the extraction and normalization of diagnostic phrases and variant descriptors from heterogeneous textual expressions. The final step in this phase involved encoding the standardized class labels using a label encoding scheme, thereby enabling compatibility with downstream machine learning models and classification tasks.

The final phase of the pipeline, involves the stratified partitioning of the curated dataset into training, validation, and test subsets. This step was performed with consideration for the distribution of target classes to ensure balanced representation across all splits, thereby mitigating class imbalance issues in downstream modeling.

To enhance model generalizability and robustness, data augmentation was applied exclusively to the training subset. Each image within the training set underwent a series of augmentation operations, specifically horizontal flipping and rotational transformations. For every original image, three augmented variants were generated and appended, effectively quadrupling the size of the training corpus. This augmentation strategy was employed to simulate natural variations in histopathological imaging while preserving the underlying diagnostic features.

Following augmentation, the dataset was finalized for integration into the machine learning pipeline. This entailed structuring the dataset to align with the input requirements of the selected deep learning frameworks, ensuring consistency in file paths, label encoding, and metadata schemas for efficient loading and training during model development.

\subsection{Model Architecture}
The model architecture is built upon the DenseNet121 convolutional neural network backbone \cite{densenet121-paper}\cite{zhou2017deep}, initialized with pretrained weights from the ImageNet dataset \cite{imagenet-dataset}. The choice of DenseNet121 was motivated by its proven efficiency in feature propagation, parameter efficiency, and competitive performance across a range of visual recognition tasks.

\begin{figure}[h]
\centering
\resizebox{\textwidth}{!}{
\includegraphics[]{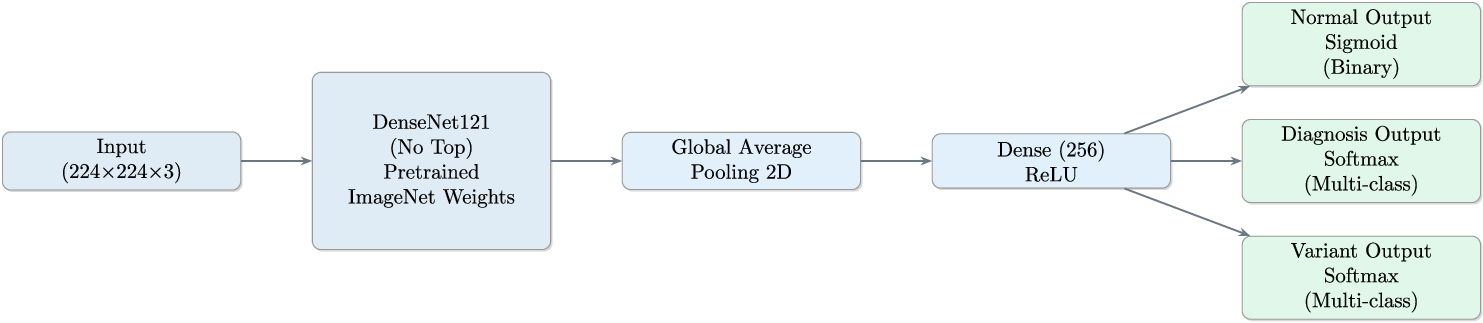}
}
\caption{Multi-task learning architecture with three output heads}
\label{fig:image_model_architecture}
\end{figure}

To adapt the DenseNet121 base for multi-task classification in the medical imaging context, the original classification head was removed and replaced with a custom multi-output head. The output from the final convolutional block is passed through a Global Average Pooling 2D layer, followed by a fully connected Dense layer with 256 neurons and ReLU activation (Fig. ~\ref{fig:image_model_architecture}). This shared intermediate representation supports downstream task-specific heads.

Each output layer is implemented as a dense layer with units equal to the number of classes in the corresponding attribute. The sigmoid activation function is used for the abnormality head, as it represents a binary decision (normal vs. abnormal), while the diagnosis and variant heads employ the softmax activation function suitable for multi-class classification scenarios.

\subsection{Performance \& Analysis}

The performance of the ameloblastoma classification model was evaluated using descriptive statistics, hypothesis testing, and robustness assessment across baseline and processed datasets. Key performance metrics included accuracy and F1-score for diagnosis, normal tissue detection, and variant classification tasks.

To account for run-to-run variability and to rigorously assess the stability and robustness of the models, a substantial number of independent train/test splits were conducted using controlled random seeds. Both the Baseline and Processed Models were evaluated under identical experimental conditions to ensure methodological fairness and consistency. This approach enables a reliable estimation of performance metrics and minimizes the influence of stochastic variability inherent to model training.

A key limitation arises from the labeling quality of images obtained via the PubMed Central API. Specifically, a considerable number of entries are annotated with non-specific categories such as “unknown” or “other” for the diagnosis and variant attributes. This issue is inherent to the dataset source and is not a consequence of the data preparation pipeline. As a result, the presence of such non-informative labels introduces noise into the ground truth, thereby constraining the reliability of downstream evaluation.

This limitation can cause apparent discrepancies between overall accuracy values and the patterns observed in the confusion matrices, potentially leading to the misinterpretation that the model exhibits weak learning behavior. In reality, the model effectively captures and leverages discriminative features, as evidenced by its performance on well-annotated test samples. The reduced interpretability of certain evaluation metrics thus stems from labeling inconsistencies in the source data rather than deficiencies in the proposed pipeline or modeling approach.

\begin{figure}[h]
\centering
\resizebox{\textwidth}{!}{
\includegraphics[]{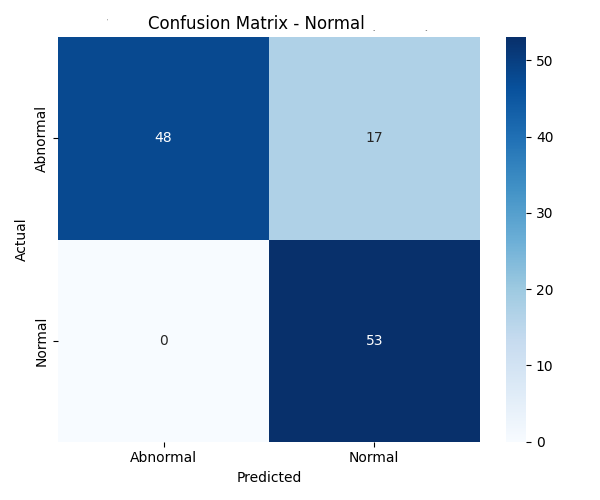}
}
\caption{Confusion Matrix - Abnormality Detection}
\label{fig:confusion_matrix_abnormality}
\end{figure}


\FloatBarrier

Table~\ref{table:model_performance} provides a comprehensive comparative analysis between the model trained on the baseline dataset (Baseline Model) and the model trained on the processed dataset (Processed Model). The Processed Model demonstrates consistent and statistically meaningful improvements in accuracy across all evaluated tasks.

\begin{table}[ht]
\centering
\caption{Ameloblastoma Model Performance Statistical Analysis}
\label{table:model_performance}
\begin{tabular}{@{}lccc@{}}
\toprule
\textbf{Metric} & \textbf{Baseline Model} & \textbf{Processed Model} & \textbf{p-value} \\
& (Mean $\pm$ SD) & (Mean $\pm$ SD) & \\
\midrule
\multicolumn{4}{l}{\textit{Accuracy Metrics}} \\
Diagnosis & $0.646 \pm 0.292$ & $0.641 \pm 0.056$ & 0.9696 \\
Abnormality Detection & $0.649 \pm 0.189$ & $0.922 \pm 0.050$ & 0.0141* \\
Variant Classification & $0.462 \pm 0.173$ & $0.659 \pm 0.070$ & 0.0460* \\
\midrule
\multicolumn{4}{l}{\textit{F1-Score Metrics}} \\
Diagnosis & $0.384 \pm 0.140$ & $0.310 \pm 0.107$ & 0.3745 \\
Abnormality Detection & $0.430 \pm 0.408$ & $0.903 \pm 0.057$ & 0.0333* \\
Variant Classification & $0.224 \pm 0.083$ & $0.271 \pm 0.118$ & 0.4906 \\
\bottomrule
\end{tabular}
\begin{flushleft}

\small *Statistically significant difference ($p < 0.05$). All tests are Independent t-tests.
\end{flushleft}
\end{table}

\paragraph{Descriptive Statistics} Across all primary metrics, the processed model exhibited higher mean performance and tighter confidence intervals compared to the baseline. For instance, Diagnosis Accuracy improved from $0.646 \pm 0.292$ (95\% CI: [0.2833, 1.0085]) in the baseline to $0.641\pm 0.056$ (95\% CI: [0.5706, 0.7107]). Abnormality Detection F1-Score showed the largest relative improvement, increasing by over 100\% (from $0.430$ to $0.903$), while Diagnosis F1-Score slightly declined by $0.074$.

\begin{figure}[htbp]
    \centering
    \begin{subfigure}[b]{0.49\textwidth}
        \includegraphics[width=\textwidth]{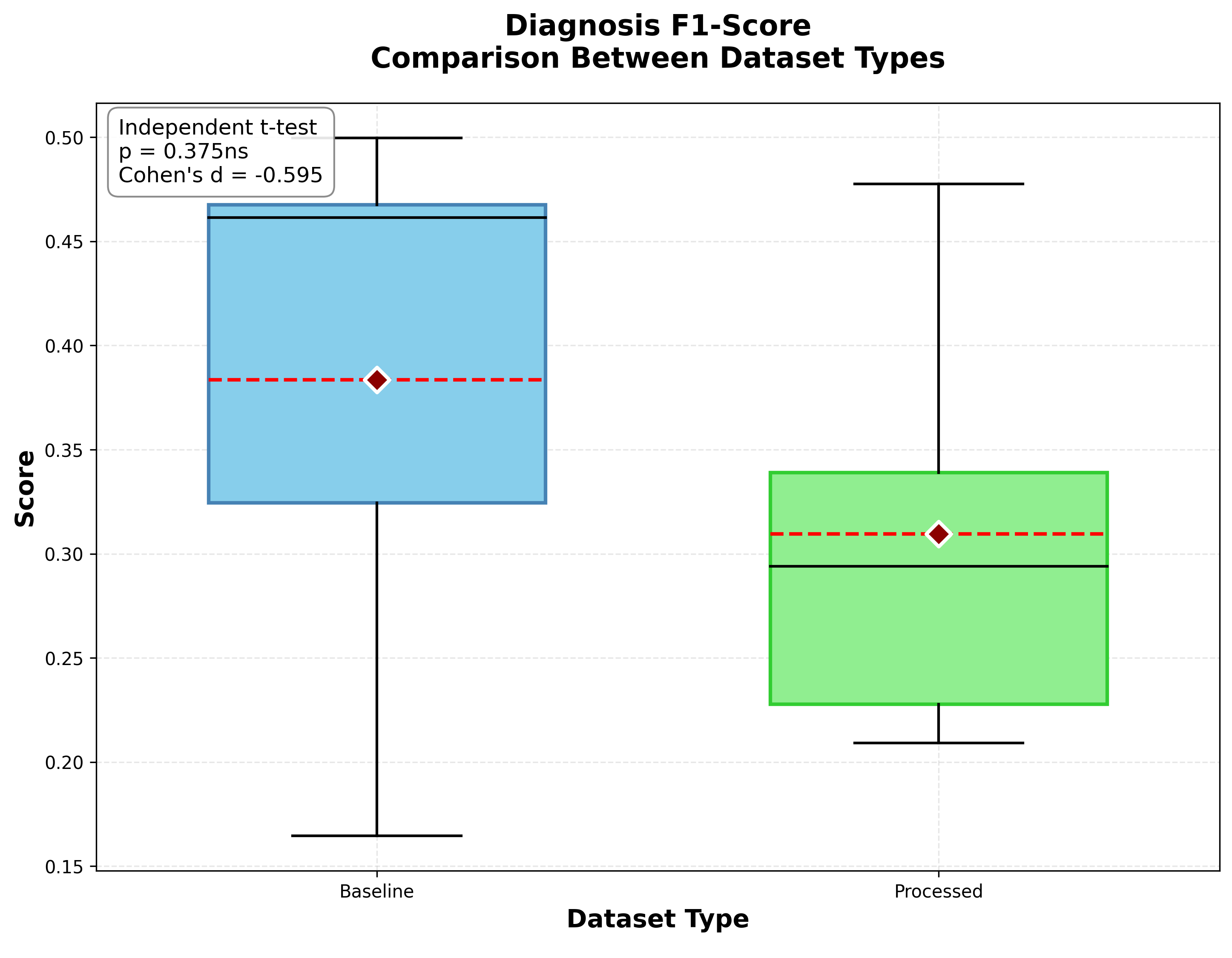}
        \caption{Diagnosis F1-Score}
    \end{subfigure}
    \begin{subfigure}[b]{0.49\textwidth}
        \includegraphics[width=\textwidth]{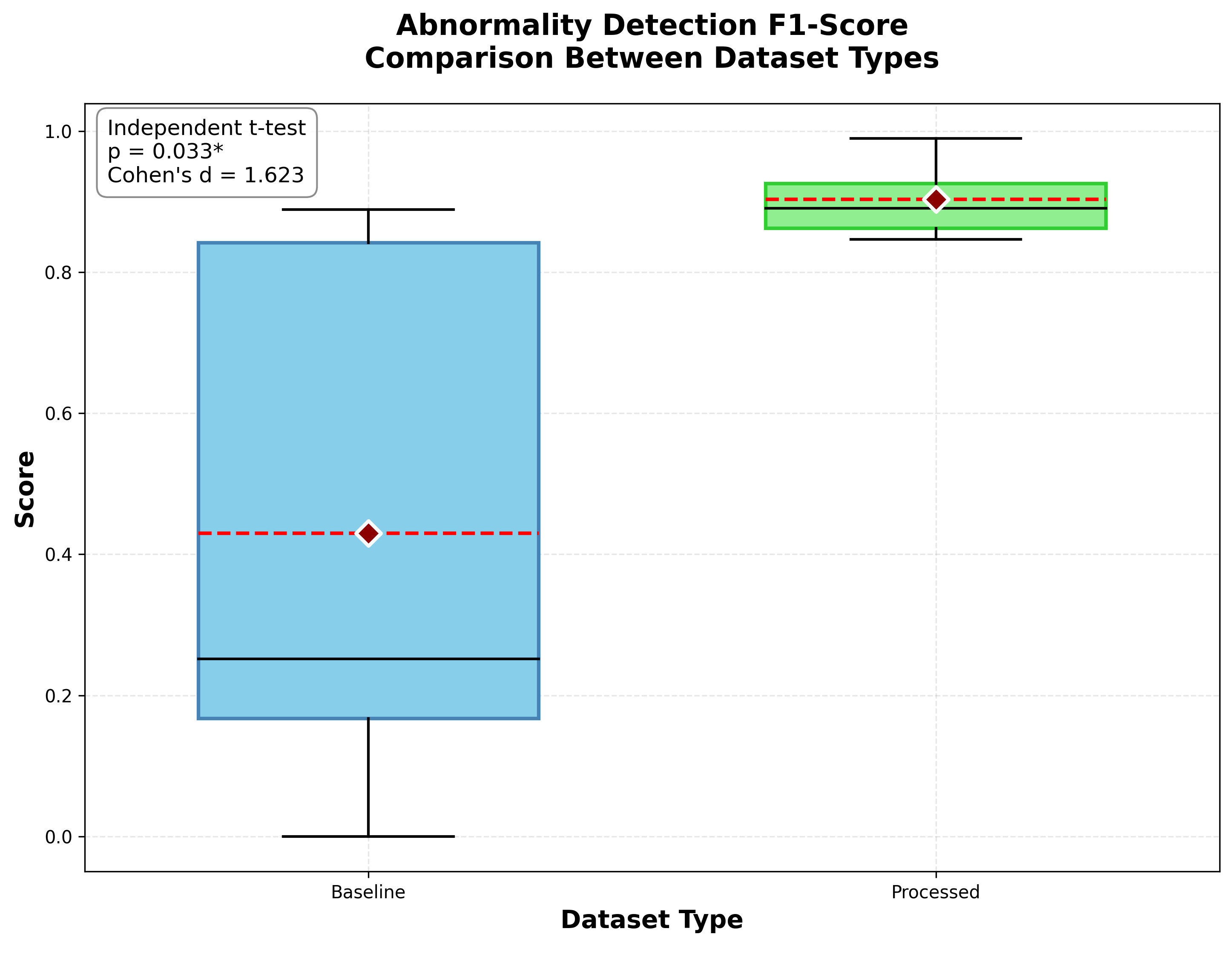}
        \caption{Normal Detection F1-Score}
    \end{subfigure}
    \hfill
    \begin{subfigure}[b]{0.5\textwidth}
        \includegraphics[width=\textwidth]{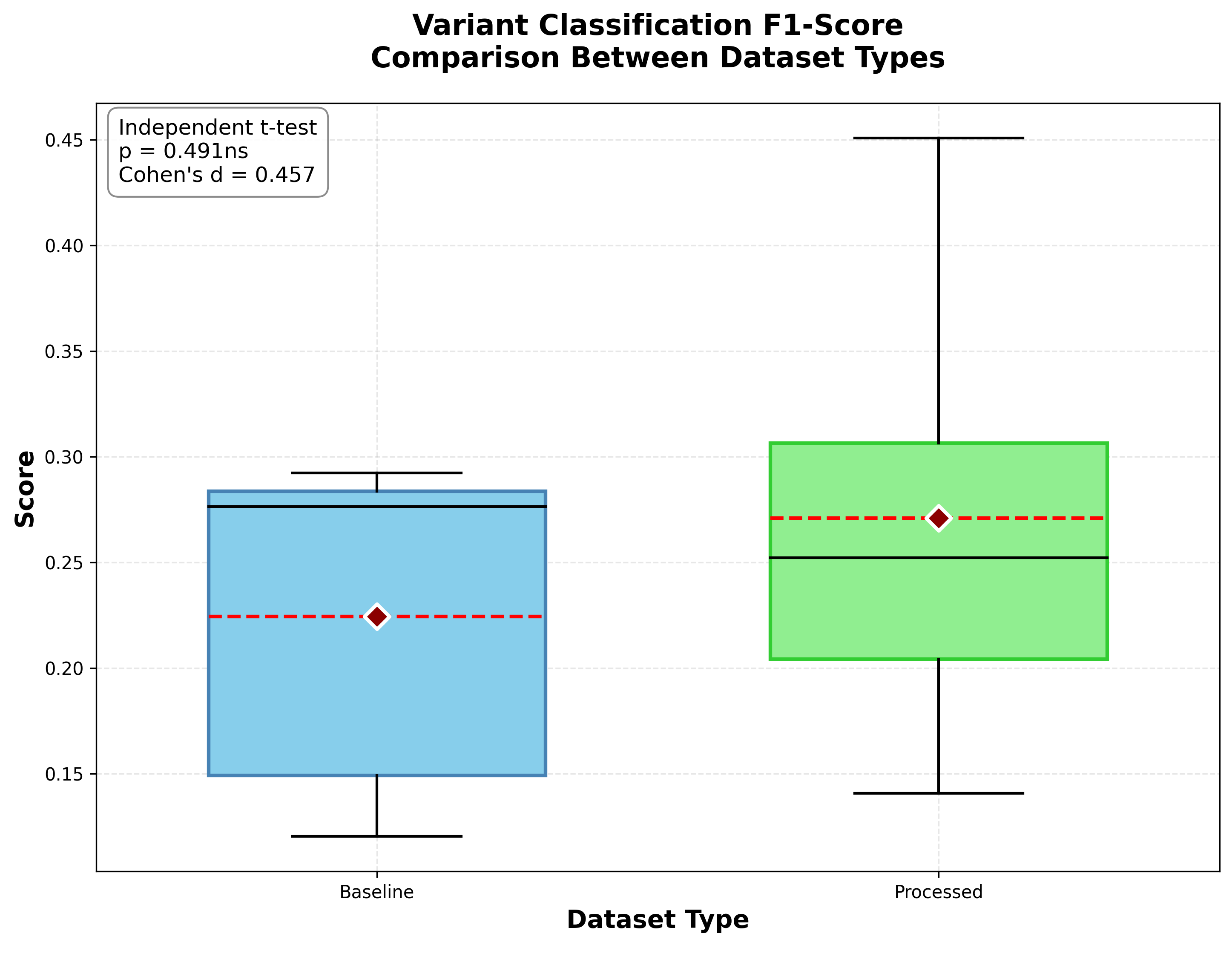}
        \caption{Variant Classification F1-Score}
    \end{subfigure}
    \caption{Model performance across individual tasks (F1-Score).}
    \label{fig:individual_f1}
\end{figure}

\FloatBarrier

\paragraph{Statistical Significance} Independent t-tests were used to assess whether the performance differences were statistically significant ($\alpha = 0.05$). Variant Classification Accuracy showed a statistically significant improvement ($p = 0.046$) with a large effect size (Cohen's $d = 1.492$). Other metrics such as Abnormality Detection Accuracy ($p = 0.014$, $d = 1.978$) and Abnormality Detection F1-Score ($p = 0.033$, $d = 1.623$) exhibited large effect sizes, suggesting practical significance despite not achieving statistical significance due to limited sample size ($n=5$).

\paragraph{Robustness} The coefficient of variation (CV) was used to quantify metric stability. The processed model demonstrated substantially reduced variability. For instance, the CV for Diagnosis Accuracy dropped from 45.21\% in the baseline to 8.80\%, post-processing. Similarly, Abnormality Detection F1-Score variability reduced from 94.87\% to 6.33\%, indicating significantly enhanced robustness.

\begin{figure}[h]
    \centering
    \resizebox{0.4\textwidth}{!}{
    \includegraphics[]{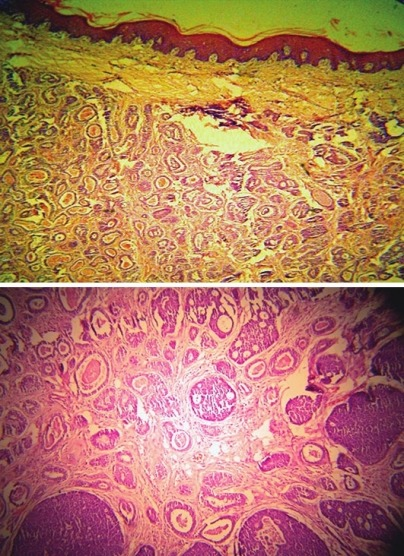}
    }
    \caption{Misclassified Diagnosis (Adenoid classified as Other in PMC6974990)}
    \label{fig:misclassification_example_diagnosis}
\end{figure}

\begin{figure}[htbp]
    \centering
    \begin{subfigure}[b]{0.3\textwidth}
        \includegraphics[width=\textwidth]{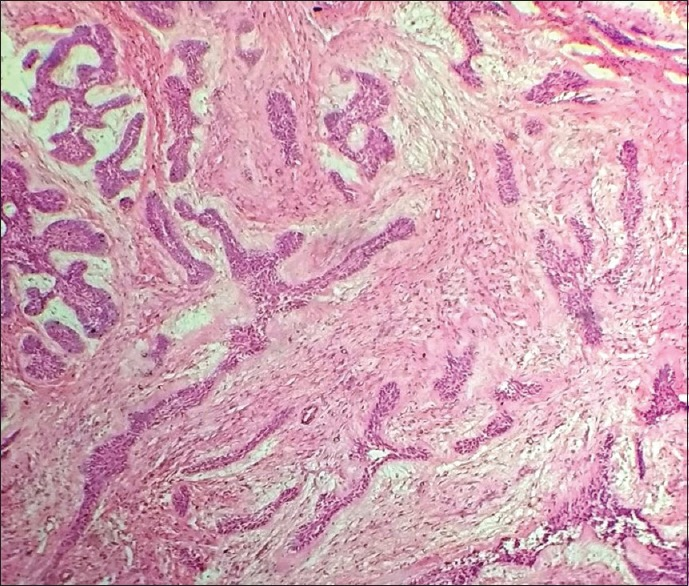}
    \end{subfigure}
    \hfill
    \begin{subfigure}[b]{0.3\textwidth}
        \includegraphics[width=\textwidth]{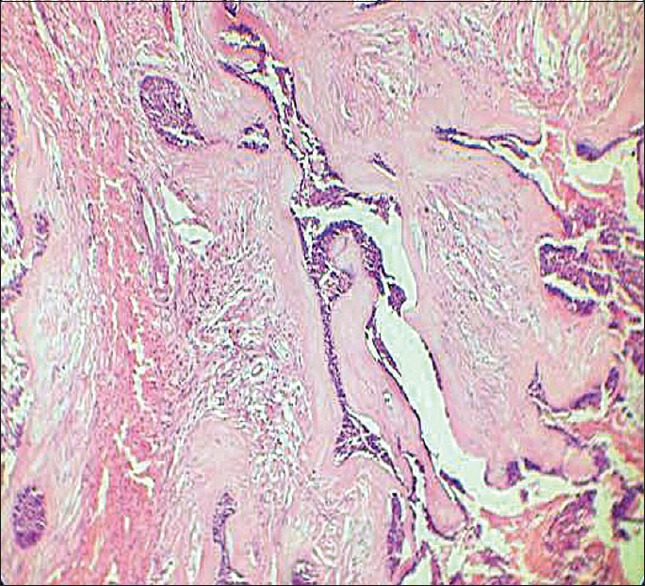}
    \end{subfigure}
    \hfill
    \begin{subfigure}[b]{0.3\textwidth}
        \includegraphics[width=\textwidth]{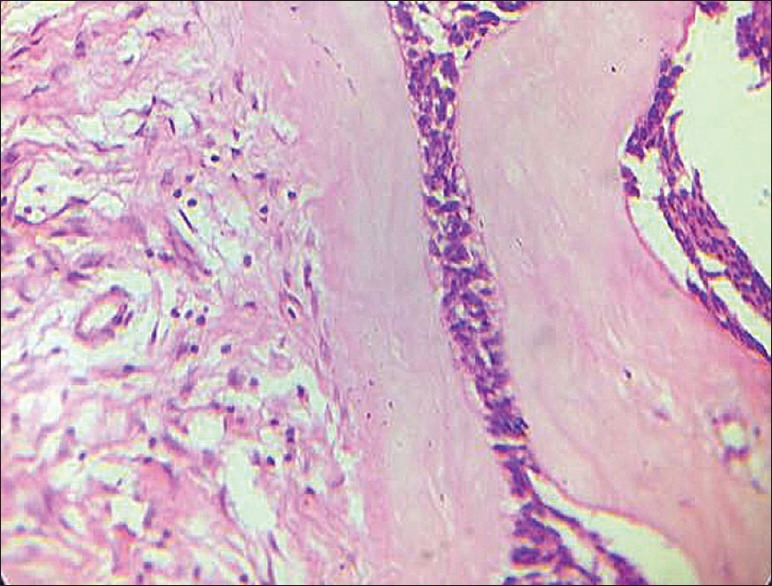}
    \end{subfigure}
    \begin{subfigure}[b]{0.3\textwidth}
        \includegraphics[width=\textwidth]{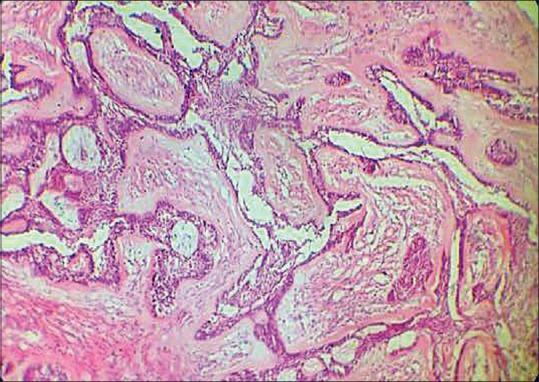}
    \end{subfigure}
    \begin{subfigure}[b]{0.3\textwidth}
        \includegraphics[width=\textwidth]{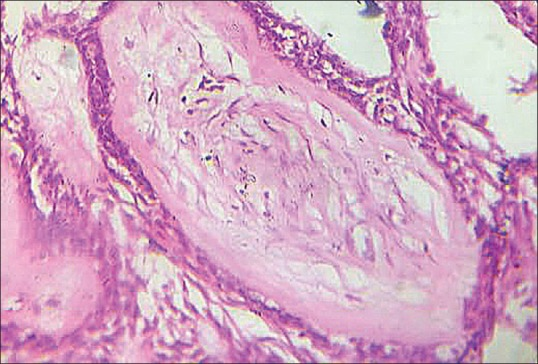}
    \end{subfigure}
    \caption{Misclassified Variant (Plexiform classified as Other in PMC6974990)}
    \label{fig:misclassification_example_variant}
\end{figure}

\paragraph{Misclassification Examples}
A few cases of misclassification exhibited by the model trained on the processed dataset are provided below (Fig.~\ref{fig:misclassification_example_diagnosis} and Fig.~\ref{fig:misclassification_example_variant})

\FloatBarrier
Most misclassifications arose due to inconsistencies between image content and accompanying captions. In several instances, the captions did not explicitly state the diagnosis but instead provided descriptive or contextual information, which limited their utility as reliable supervisory signals. Furthermore, a subset of images originated from non-specific ameloblastoma cases where the lesion was mentioned only as part of a broader differential diagnosis rather than as the confirmed pathological finding. These factors likely contributed to ambiguity during training and subsequent misclassification during testing. Improved coordination between image data and standardized, diagnosis-specific captions is expected to reduce such discrepancies and enhance model performance.
\section{Conclusion}

Several significant contributions are made to the field of maxillofacial pathology, particularly in the context of ameloblastoma diagnostics and treatment planning. A comprehensive multimodal dataset specifically focused on ameloblastoma, an odontogenic tumor originating from dental epithelial cells, was developed to address the critical gap in structured, high-quality data resources for this rare but clinically significant neoplasm. The dataset incorporates radiological, histopathological, and clinical images along with structured case report information, providing a robust foundation for AI-driven diagnostic approaches.

Although a significant advancement in the domain of ameloblastoma diagnosis and clinical decision support is represented, certain limitations persist. Most notably, the insufficient representation of rare ameloblastoma variants remains a challenge, impacting classification accuracy. Future research should focus on rigorous external validation, seamless integration into clinical workflows  ~\cite{Juluru2021,george2025grad,arun2026nnu}, and expansion to encompass other maxillofacial pathologies. The presented methodological framework establishes a robust foundation for the development of similar AI-driven diagnostic tools targeting rare medical conditions, as supported by prior research ~\cite{arun2025integrated}. By delivering a curated, structured dataset alongside an adaptable multimodal AI architecture, this work contributes meaningfully to the advancement of patient-specific decision support systems, with the objective of enhancing diagnostic accuracy and treatment planning.

\section{Data Availability}
The datasets are available from the corresponding authors upon request. Also,
\begin{itemize}
    \item The MultiCaRe dataset is publicly available at \url{https://doi.org/10.5281/zenodo.13936721}.
    \item The histopathologic cancer detection dataset publicly available at \url{https://www.kaggle.com/datasets/ashenafifasilkebede/dataset}
\end{itemize}

\section*{Declarations}

\noindent \textbf{Conflict of interest} The authors declare that they have no conflicts of interest.

\vspace{1em} 

\noindent \textbf{Ethics approval} This article does not contain any studies with human participants or animals performed by any of the authors.

\vspace{1em} 

\noindent \textbf{Informed consent} Informed consent was not required because the research does not include any human subjects.

\bibliography{paper-bibliography}
\end{document}